\documentclass[pdflatex,sn-mathphys-num]{sn-jnl} 
\usepackage{siunitx}
\usepackage{pdflscape}
\usepackage{longtable}
\usepackage{rotating}
\usepackage{graphicx}%
\usepackage{makecell}
\usepackage{multirow}%
\usepackage{array}
\usepackage{lipsum}  
\usepackage{amsmath}
\usepackage{amssymb}
\usepackage{mathrsfs}%
\usepackage{pdflscape}   
\usepackage{longtable}
\usepackage{xcolor}%
\usepackage{textcomp}%
\usepackage{manyfoot}%
\usepackage{booktabs}%
\usepackage{algorithm}%
\usepackage{algorithmicx}%
\usepackage{algpseudocode}%
\usepackage{caption}
\usepackage{colortbl}
\usepackage[table]{xcolor}%
\usepackage{listings}%
\usepackage{float}  % add this in your preamble if not already present

\theoremstyle{thmstyleone}%
\theoremstyle{thmstyletwo}%

\theoremstyle{thmstylethree}%

\usepackage{array}
\usepackage{colortbl}
\usepackage{hhline}
\usepackage{booktabs}
\usepackage{multirow}
\usepackage{booktabs}
\usepackage[table]{xcolor} % Optional, for better formatting

\begin{document}

\title[A Survey of the State-of-the-Art in Conversational Question Answering Systems]{A Survey of the State-of-the-Art in Conversational Question Answering Systems}

\author[1]{\fnm{Manoj Madushanka} \sur{Perera}}\email{manoj.perera@students.mq.edu.au}

\author*[1]{\fnm{Adnan} \sur{Mahmood}}\email{adnan.mahmood@mq.edu.au}

\author[1]{\fnm{Kasun Eranda} \sur{Wijethilake}}\email{kasun.wijethilake@mq.edu.au}

\author[1]{\fnm{Fahmida} \sur{Islam}}\email{fahmida.islam@mq.edu.au}

\author[1]{\fnm{Maryam} \sur{Tahermazandarani}}\email{maryam.tahermazandarani@mq.edu.au}

\author[1]{\fnm{Quan Z.} \sur{Sheng}}\email{michael.sheng@mq.edu.au}

\affil[1]{\orgdiv{School of Computing}, \orgname{Macquarie University}, \city{Sydney}, \state{NSW} \postcode{2109} \country{Australia}}

\abstract{Conversational Question Answering (ConvQA) systems have emerged as a pivotal area within Natural Language Processing (NLP) by driving advancements that enable machines to engage in dynamic and context-aware conversations. These capabilities are increasingly being applied across various domains, i.e., customer support, education, legal, and healthcare where maintaining a coherent and relevant conversation is essential. Building on recent advancements, this survey provides a comprehensive analysis of the state-of-the-art in ConvQA. This survey begins by examining the core components of ConvQA systems, i.e., history selection, question understanding, and answer prediction, highlighting their interplay in ensuring coherence and relevance in multi-turn conversations. It further investigates the use of advanced machine learning techniques, including but not limited to, reinforcement learning, contrastive learning, and transfer learning to improve ConvQA accuracy and efficiency. The pivotal role of large language models, i.e., RoBERTa, GPT-4, Gemini 2.0 Flash, Mistral 7B, and LLaMA 3, is also explored, thereby showcasing their impact through data scalability and architectural advancements. Additionally, this survey presents a comprehensive analysis of key ConvQA datasets and concludes by outlining open research directions. Overall, this work offers a comprehensive overview of the ConvQA landscape and provides valuable insights to guide future advancements in the field.}


\maketitle

\section{Introduction}

Natural Language Processing (NLP) is a promising interdisciplinary paradigm that encompasses computational linguistics, Artificial Intelligence (AI), and data science with the primary focus being the interaction between computers and human languages \cite{orynbay2025role,frenda2024perspectivist}. It integrates rule-based computational methods with statistical and machine learning approaches. This thus enables machines to interpret, comprehend, and produce human text and speech \cite{Nagarhalli2021ImpactReview}. In the early days of NLP, i.e., during the 1940s -- 1950s, Warren Weaver, an American scientist and mathematician, proposed the idea of computer-based language translation that focuses on translating text word by word \cite{melby2019future}. However, this approach faced significant challenges in accurately translating the semantics and grammar of languages \cite{Jiang2020NaturalReview}.  This led to a divide in NLP research between two groups --  \textit{symbolists} and \textit{frequentists}. Symbolists focused on deep language analysis, whereas, frequentists employed statistical techniques. During the 1960s -- 1970s, there were substantial advances in machine translation and semantic analysis. This period was notable for the application of logic methods in NLP. 
Additionally, this era saw the creation of influential systems, i.e., Prolog \cite{Gupta2023} and SHRDLU \cite{Johnson1988, SHRDU}. During the 1970s -- 1980s, NLP faced a period of stagnation owing to inadequate computing power \cite{AIStagnation}. Nevertheless, NLP has advanced considerably since the 1990s and onwards \cite{1990}. This period has been characterized by large-scale computational power and enhanced usability. Large text corpora and major developments in computer science and statistics are the main drivers of recent advances.

\begin{figure}[!t]
    \centering
    \includegraphics[width=1.0\textwidth]{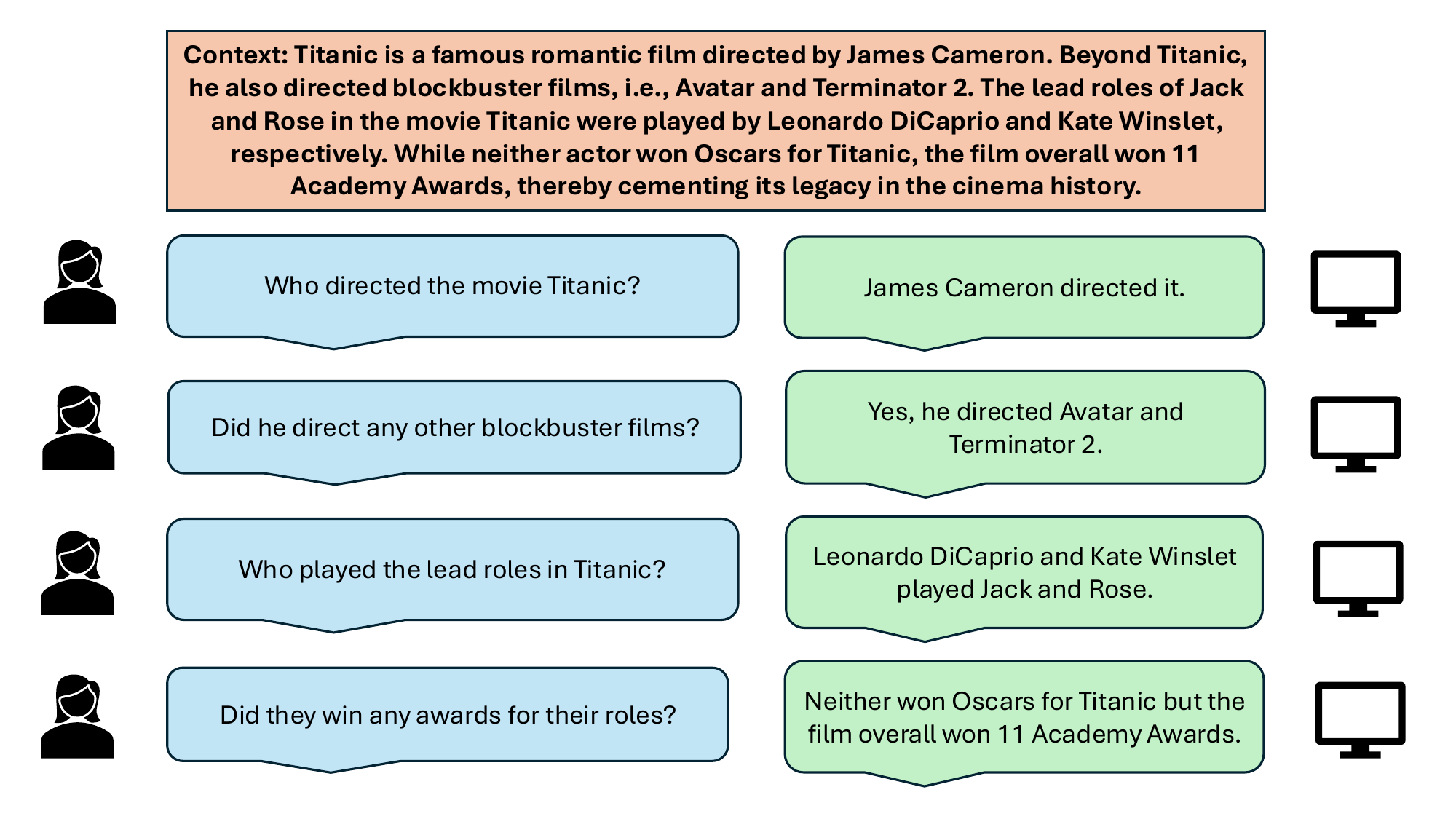} % Adjust the width as needed
    \vspace{-10pt} % Reduce the space between the image and caption
    \captionsetup{justification=centering}
    \caption{Illustration of a conversational question answering interaction showcasing how contextual understanding and follow-up questions enhance natural conversation flow in real-world scenarios.}
    \label{fig:cqa_modeEx}
\end{figure}

Building on the fundamental advances in NLP, a wide range of language-based techniques have been developed. These techniques include tokenization \cite{Tokenization}, part-of-speech tagging \cite{partofspeech}, syntax parsing \cite{syntax}, sentiment analysis \cite{sentiment}, word embeddings \cite{wordembeddings}, machine translation \cite{machinetranslation}, Question Answering (QA) \cite{qafornlp}, and deep learning \cite{Rayhan2023NATURALLANGUAGE, deeplearnig}. Such developments made it easier for machines to manage and use a huge volume of textual content efficiently. Within the realm of NLP, QA stands as a specialized subset \cite{bai2025ramqaunifiedframeworkretrievalaugmented}. It focuses on understanding, generating, and representing natural language. This enables a system to automatically retrieve relevant answers from a knowledge base for any given question \cite{Abdel-Nabi2023DeepSurvey}. A further refinement within QA is Conversational Question Answering (ConvQA). This advanced technique demands an extensive understanding of conversational context which extends beyond the boundaries of standard single-turn Machine Reading Comprehension (MRC) tasks \cite{KTurn2}. Figure \ref{fig:cqa_modeEx} illustrates a sample interaction in ConvQA depicting how users and systems engage in multi-turn conversations to answer queries in context.

\begin{figure}[!t]
    \centering
    \includegraphics[width=1.0\textwidth]{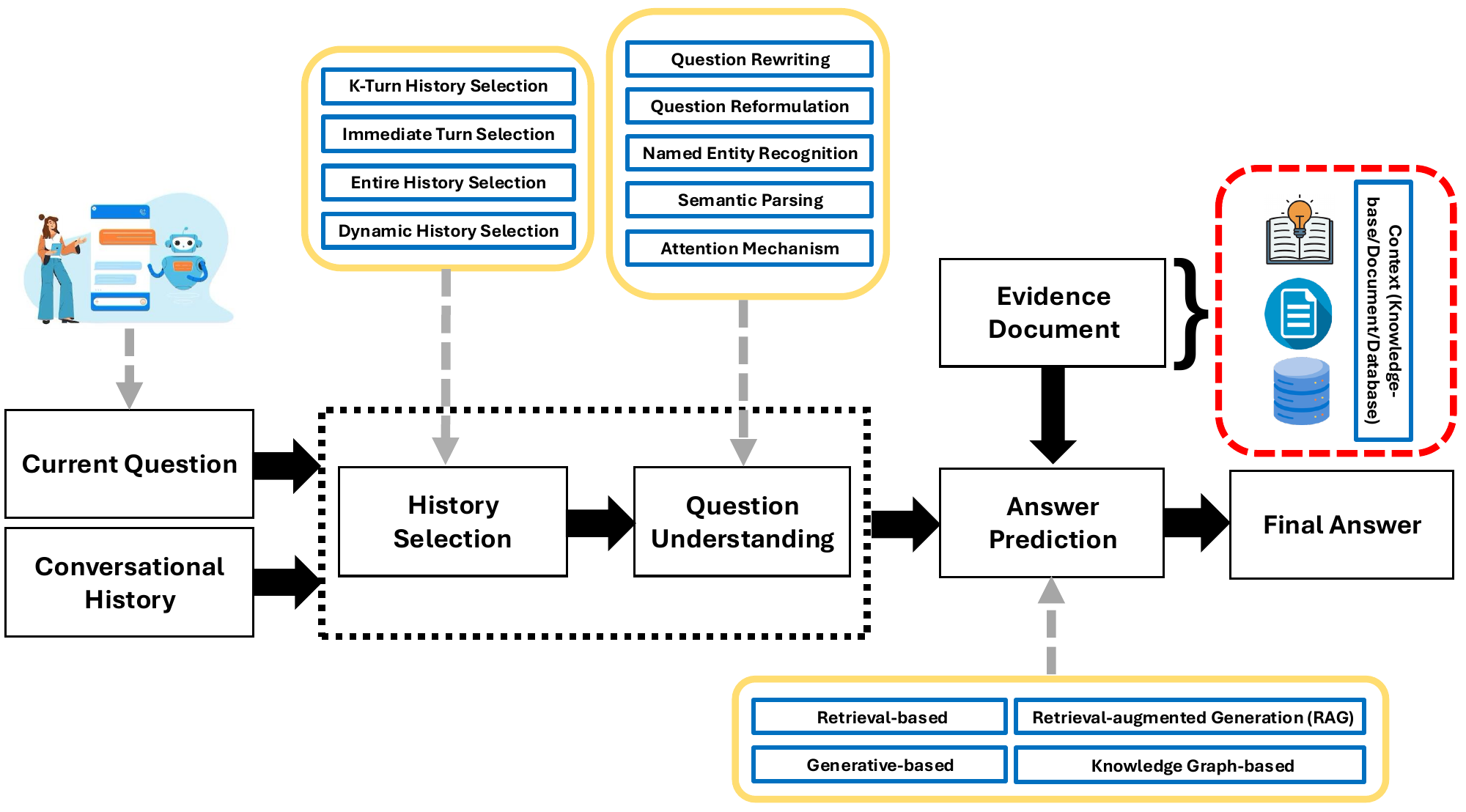} % Adjust the width as needed
    \vspace{-5pt} % Reduce the space between the image and caption
    \captionsetup{justification=centering}
    \caption{Architecture of a conversational question answering system.}
    \label{fig:cqa_model}
\end{figure}

ConvQA systems enable human-like interactions across a wide range of applications. These systems are tasked with handling multi-turn conversations, thereby requiring sophisticated techniques to maintain coherence and deliver accurate responses. Figure~\ref{fig:cqa_model} delineates the architecture of a typical ConvQA system and demonstrates how various key components, i.e., history selection, question understanding, and answer prediction, work together to process users' questions to generate more accurate answers. The current question is analyzed vis-\`a-vis
the conversational history, thereby allowing the ConvQA system to select relevant portions of the historical conversational turns for effective question understanding. The evidence document or the context paragraph is then used to extract or generate accurate answers for a user's question. This process ensures that the ConvQA system maintains coherence and relevance throughout the multi-turn interactions.

ConvQA systems aim to make interactions with Large Language Models (LLMs) more natural and intuitive \cite{CQASurvey,2024_05}, and can be divided into three types -- (a) \textit{task-oriented dialog systems} for executing particular tasks, e.g., booking reservations, (b) \textit{chat-oriented dialog systems} to perform human interactions, e.g., ChatGPT and Google Assistant, and (c) \textit{QA dialog systems} for providing brief answers based on diverse data sources \cite{Gao2018NeuralAI}, e.g., medical QA systems. ConvQA techniques form the foundation of QA dialog systems. It enables machines to answer questions based on certain passages.  This capability has the potential to substantially change the way humans interact with machines. If users request more information, this interaction may transform into a multi-turn conversation \cite{ImmediateTurn2}.

\begin{figure}[!t]
    \centering
    \includegraphics[width=1.0\textwidth]{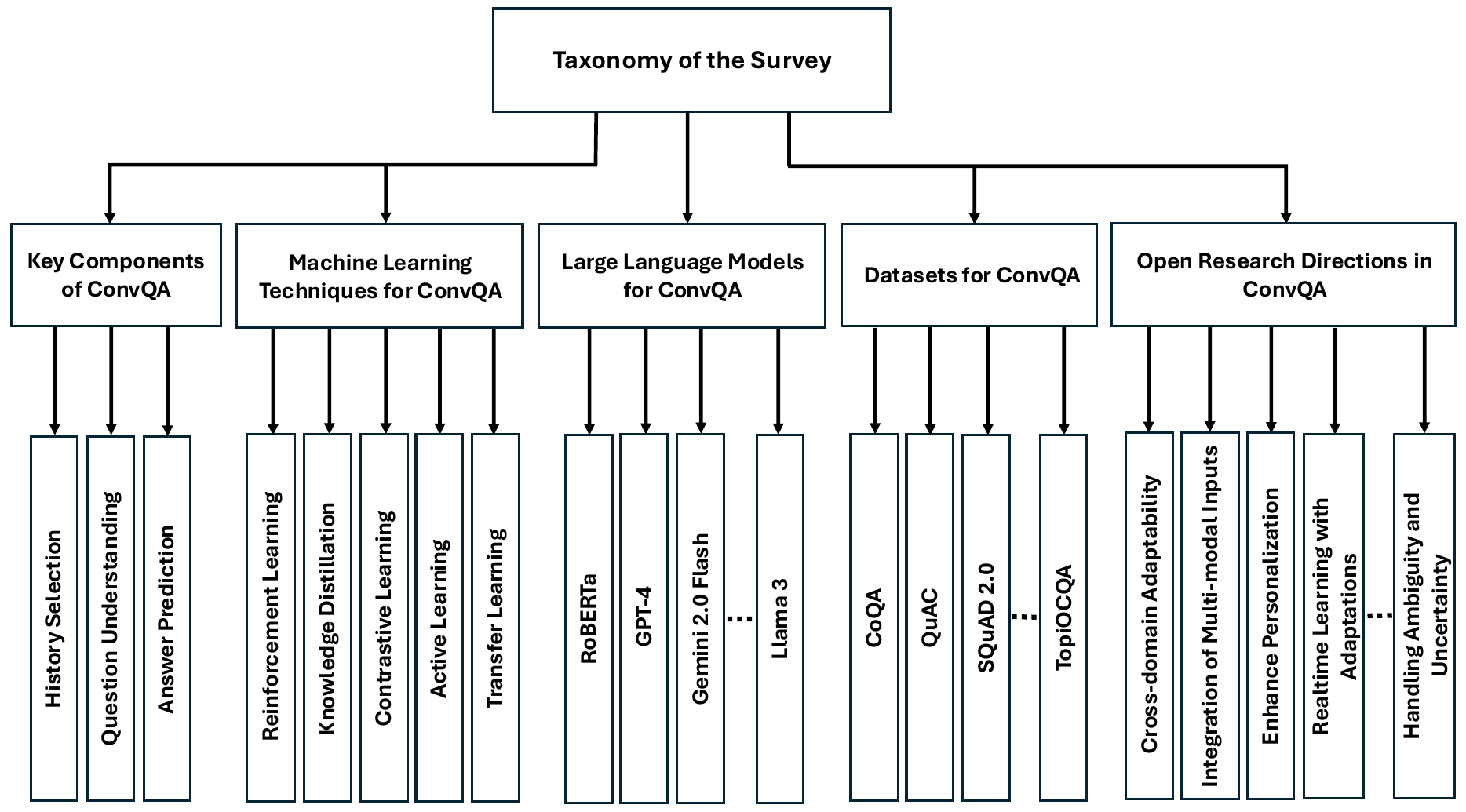} % Adjust the width as needed
    \vspace{-5pt} % Reduce the space between the image and caption
    \captionsetup{justification=centering}
    \caption{Taxonomy of the survey.}
    \label{fig:taxanomy}
\end{figure}

As delineated in Figure~\ref{fig:taxanomy}, this survey explores several critical aspects of the ConvQA systems. First, we examine the key components including history selection, question understanding, and answer prediction which are indispensable for maintaining coherence in multi-turn conversations. Next, we delve into advanced machine learning techniques that enhance ConvQA performance. Furthermore, we analyze LLMs and datasets that shape ConvQA research and applications. Finally, we highlight open research directions that will guide the future development of more robust and adaptive ConvQA systems.

\subsection{Paper Selection Strategy} 

Primarily, Google Scholar was utilized to identify state-of-the-art research literature. Apart from the same, IEEE Xplore, ACM Digital Library, SpringerLink, ScienceDirect, and Papers with Codes were utilized. Keywords pertinent to NLP and ConvQA were specifically employed to extract relevant research literature from the said sources. This survey thus covers research literature from (a) highly prestigious journals, including but not limited to, Artificial Intelligence Review, Expert Systems with Applications, IEEE Access, Knowledge and Information Systems (KAIS), ACM Computing Surveys, IEEE Computational Intelligence Magazine and (b) renowned international conferences, including but not limited to, Association for Computational Linguistics	(ACL), North American Chapter of the Association for Computational Linguistics (NAACL), Conference on Empirical Methods in Natural Language Processing (EMNLP), International Conference on Learning Representations (ICLR), Association for the Advancement of Artificial Intelligence Conference (AAAI), International Joint Conference on Artificial Intelligence (IJCAI), Conference on Information and Knowledge Management (CIKM), ACM SIGKDD Conference on Knowledge Discovery and Data Mining	(SIGKDD), and ACM International Conference on Web Search and Data Mining	(WSDM). In addition to journal papers and conference proceedings, the search for relevant research literature was extended by incorporating preprint repositories, e.g., arXiv. These sources frequently provide early versions of the research papers, i.e., which are often either awaiting peer review or have been reviewed and are awaiting final approval or are in the production stage. This dual source approach has provided a robust representation of the state-of-the-art research.

All of the gathered research literature was subsequently organized and ranked according to diverse criteria, i.e., the quality of the research and the significant impact of the research on the domain. The shortlisted literature, therefore, offers insights into the most significant and transformative works in the field, thereby driving the future of ConvQA systems and their applications. Moreover, the citation count was taken into consideration as a measure of a paper's significance and acceptability within the academic community as a whole. 

\subsection{Key Contributions of the Survey}

This survey makes several important contributions to the field of ConvQA and which are delineated as follows:

\begin{enumerate}
    \item The survey provides a detailed analysis of the key components of ConvQA systems by focusing on history selection, question understanding, and answer prediction. By examining these components, the survey highlights how recent advancements improve the coherence and relevance of multi-turn conversations which are imperative for enhancing ConvQA performance.

    \item The survey delves into the integration of advanced machine learning techniques, i.e., reinforcement learning, knowledge distillation, contrastive learning, active learning, and transfer learning. It demonstrates how these techniques collectively enhance the efficacy, accuracy, and adaptability of ConvQA models.

    \item The survey presents a dedicated analysis of LLMs for ConvQA by examining state-of-the-art models, e.g., RoBERTa, GPT-4, Gemini 2.0 Flash, Mistral 7B, and LLaMA 3. It highlights their unique characteristics in terms of parameter size, context window length, training tokens and training data. This contribution underscores how these LLMs have advanced ConvQA capabilities through improved reasoning, extended context handling, and domain adaptability.

    \item The survey dedicates a significant section to open research directions by proposing new avenues for improving ConvQA systems. It addresses emerging challenges, i.e., cross-domain adaptability, integration of multimodal inputs, enhanced personalization, real-time learning with adaptation, dynamic conversational history management, and handling ambiguity and uncertainty. These areas are essential for pushing the boundaries of the current ConvQA systems and making them more robust, contextually aware, and user-centric.

\end{enumerate}

Through these contributions, our survey aims to enhance the understanding, development, and application of ConvQA systems for supporting ongoing research and innovation in this dynamic field.

\section{Key Components of Conversational Question Answering Systems}

\begin{figure}[!t]
    \centering
    \includegraphics[width=0.90\textwidth]{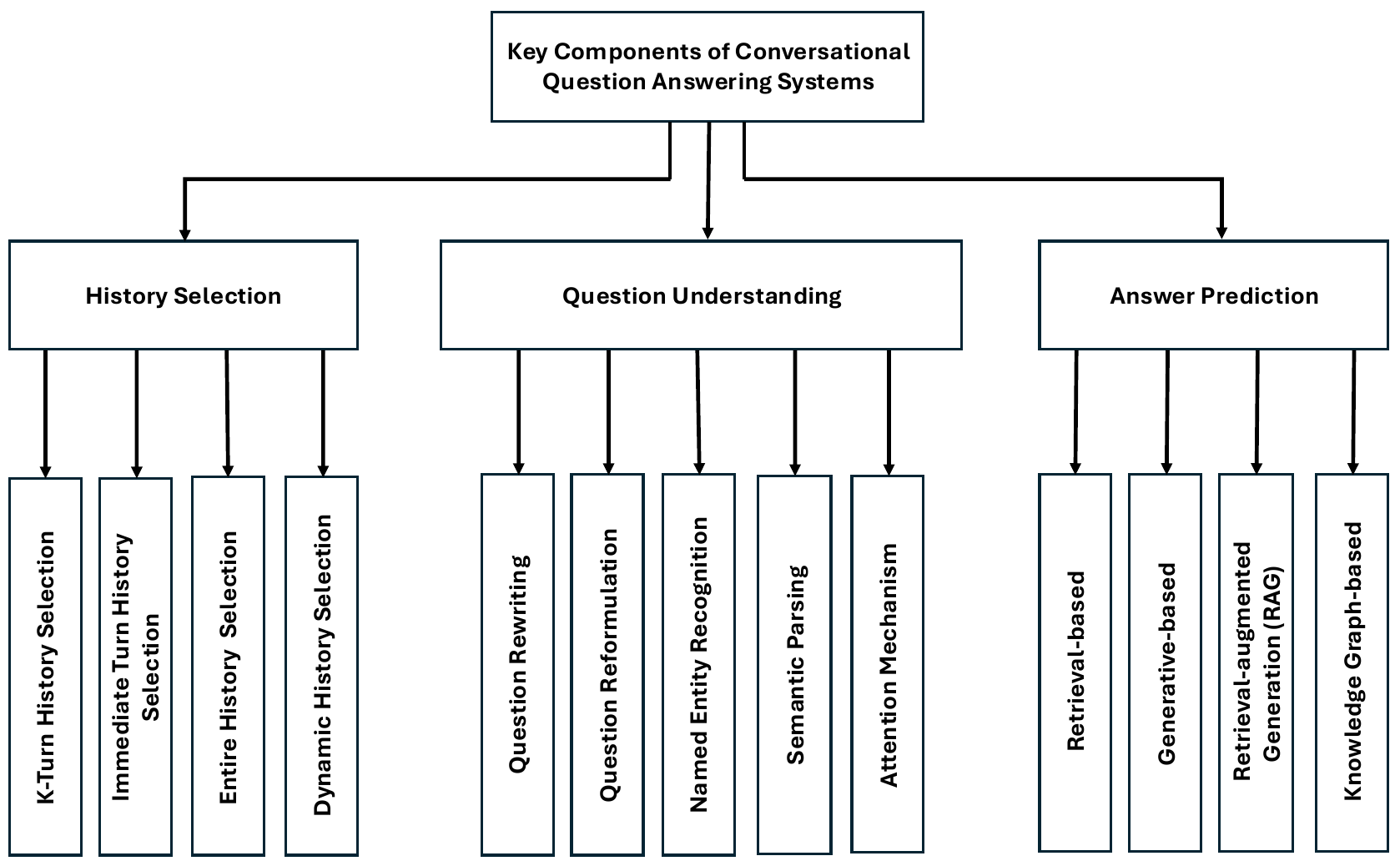} % Adjust the width as needed
    \vspace{10pt} % Reduce the space between the image and caption
    \captionsetup{justification=centering}
    \caption{Key components of conversational question answering systems.}
    \label{fig:cqa_components}
\end{figure}

ConvQA systems have complex architectures which are composed of several interdependent components as delineated in Figure~\ref{fig:cqa_components}. Each component plays a crucial role in understanding and responding to users' questions. These components improve a ConvQA system's ability to manage conversational context, interpret natural language and produce contextually accurate responses. Understanding how each component functions individually and collaboratively is essential for developing effective ConvQA solutions that can generalize across diverse conversational scenarios.

%Furthermore, Table \ref{comparison_table} delineates how existing literature has utilized these key components.

\subsection{History Selection}
Effective history selection allows ConvQA systems to maintain continuous flow of a conversation by keeping track of the information shared between a user and a system \cite{CQASurvey,CQAHistory1,article11}. The history selection process involves understanding the context of the current question and retaining the optimal conversation history to generate accurate responses. History selection is crucial for ConvQA systems as it helps maintain more cohesive and relevant user interactions across multiple conversational turns. However, while the relevant historical conversational turns are crucial, irrelevant conversational turns can introduce noise \cite{Zaib2021BERT-CoQAC:Context, Tian2017HowModels}. Therefore, selecting relevant historical conversational turns is of the utmost essence and can be achieved through several approaches as outlined below.

\subsubsection{$K$-Turn History Selection} 

This is a static and simple approach, wherein, a predetermined number of $K$ preceding conversational turns are taken into account for generating or retrieving an answer to a user's question. Whilst this method is quite simple to apply, it poses inherent risks. The biggest challenge lies in the determination of an optimal value of $K$. The ConvQA systems may generate incomplete or inaccurate responses if $K$ is insufficiently small. In contrast, a ConvQA system may incorporate irrelevant details from earlier parts of the conversation if $K$ is excessively large, thereby introducing noise and reducing the relevance of an answer \cite{EntireHistory1, 2024_03}. $K$-turn history selection was utilized in \cite{KTurn2} by prepending the current question with the latest N rounds of previous conversational turns, thereby effectively converting the task into a standard machine reading comprehension problem.  Through ablation studies, the authors discovered that using the last 2 turns ($K$=2) yielded the best performance on the CoQA dataset. In addition, $K$-turn history selection was envisaged in \cite{KTurn1}, where the authors incorporated the 5 most recent conversational turns as input to their question rewriting model. These turns were concatenated using a special [SEP] token to preserve turn boundaries. This fixed-length context enabled the model to disambiguate follow-up questions by generating explicit question rewrites, thereby facilitating effective downstream QA. Furthermore, the $K$-turn history selection approach was employed in \cite{ImmediateTurn1}, where the model incorporates the current question along with the previous $K$ turns into the BERT encoder. Experiments showed that utilizing $K$ = 2 turns gave the best performance on QuAC, thereby highlighting the benefit of a limited but focused conversational history for context modeling. Furthermore, $K$-turn history selection approach was utilized in \cite{ImmediateTurn2}, where the model selected a fixed number of recent conversational turns. This rule-based strategy assumed that more recent conversational turns are more relevant, and experiments showed that selecting around 5–6 previous turns yielded the best performance.

\subsubsection{Immediate Turn Selection}

In ConvQA systems, leveraging the immediate historical conversational turns involves utilizing only the most recent interaction(s) within the conversation. This approach focuses on incorporating the most recent one or two conversational turns to maintain a concise and contextually relevant understanding of the current user question. The primary advantage of this method is its ability to quickly adapt to the current conversational context, thereby enabling the model to provide accurate and contextually appropriate responses without being overwhelmed by an extensive conversational history. Immediate turn history selection approach was utilized in \cite{KG2}, where only the previous question, its corresponding answer, and the current question were considered as the input to the QA model in the ConvQA system. These conversational turns were separated using a [SEP] token and fed into a transformer encoder. This setup enabled the QA model to resolve coreference and ellipsis effectively without relying on the entire conversational history. Additionally, the immediate turn  history selection approach was incorporated in \cite{semantic1} by using only the immediate previous conversational turn as the conversational context when parsing the current user question. This limited context is concatenated with the current question and input into a BERT-based encoder to predict the corresponding SPARQL query. This approach enables the model to resolve recent coreference and ellipsis while maintaining computational efficiency. Furthermore, the same history selection approach was utilized in \cite{Robust} by including only the first and immediately preceding conversational turns of the conversational history as input to the reformulation generator.  Specifically, the input was formed by concatenating the initial question-answer pair, followed by all prior question-answer pairs up to the one immediately before the current conversational turn, along with the current question and a reformulation category tag. This allowed the model to generate intent-preserving reformulations using minimal but relevant context. Moreover, the immediate turn history selection approach was envisaged in \cite{2025_2}, where only the most recent prior question from the conversational history was included during policy decomposition. This limited conversational history is used to guide the generation of relevant follow-up questions and to construct logical formulae involving both current and past conversational turns. By integrating the immediate conversational history into logical reasoning, the system enhances its ability to transparently identify unanswered conditions and determine whether additional information is required, thereby improving the explainability, precision, and decision-making flow of the overall policy compliance detection framework.

\subsubsection{Entire Conversational History Selection}

Selecting the entire conversational history in ConvQA systems involves considering all previous interactions within a conversation to maintain a comprehensive context. This approach aims to capture the entire conversational scope, thereby allowing the model to reference information from any part of the conversational history. While this approach can lead to a more thorough understanding of user questions, it also poses challenges, i.e., increased computational complexity and potential performance degradation due to the overwhelming amount of context. The entire conversational history was incorporated in \cite{EntireHistory2}, wherein the conversational history up to the previous conversational turn was used along with the current question in a reinforcement learning-based question rewriting setup. This full context is reformulated into a self-contained question which was then used by the QA model for answer prediction. Additionally, entire conversational history was incorporated in \cite{RAG1Correct}, where the full preceding conversational history was used along with the current question for both question reformulation and response generation. The reformulation component explicitly clarified ambiguous terms based on earlier conversational turns. This approach enabled better context grounding and improved the relevance of retrieved evidence, thereby enhancing the overall answer accuracy in conversational settings. Moreover, the entire conversational history was utilized in a more dynamic manner using text summarization and entity extraction techniques when responding to the current user question in \cite{MYACM}. Furthermore, entire conversational history was incorporated in \cite{EXCORD} to model the full conversational context for each question by including all previous question-answer pairs up to the current conversational turn, thereby ensuring that dependencies, i.e., anaphora and ellipsis, are resolved using the complete conversational history. This helps the model better understand context-dependent questions and improve answer consistency. Furthermore, the entire history selection approach was envisaged in \cite{2025_1}, wherein all previous question-answer pairs are considered when reformulating the current question. This comprehensive history is passed to the question understanding module to generate improved reformulations that guide better evidence retrieval and answer generation. By leveraging the full conversational context, their proposed framework enhances relevance and factuality in downstream stages.

\subsubsection{Dynamic History Selection} 
Dynamic history selection refers to adapting and utilizing relevant parts of the conversational history to maintain context and generate appropriate responses. Unlike static approaches, dynamic history selection involves real-time analysis and decision-making to determine which parts of the conversational history are pertinent to the current question. This adaptability is crucial for handling complex and extended conversations. The dynamic history selection can be further divided into two categories -- \textit{hard history selection} and \textit{soft history selection} \cite{Zaib2023KeepingAnswering}. \textit{Hard history selection} involves explicitly choosing specific conversational turns from the conversational history when generating responses. This approach makes a binary decision by including certain parts of conversational history while other parts are excluded. \textit{Soft history selection} involves assigning different weights to various parts of the conversational history, thereby allowing the ConvQA system to consider all historical conversational turns but with varying degrees of importance. This method leverages mechanisms, i.e., attention to dynamically assess and utilize the context. Reinforcement learning was employed in \cite{Qiu2021ReinforcedAnswering} to dynamically backtrack and select contextually relevant conversational history turns, thereby enhancing the ConvQA system’s understanding of the current question. A new framework entitled, Dynamic History Selection in Conversational Question Answering (DHS-ConvQA), was proposed in \cite{ZaibLearningtoselect} to dynamically select relevant conversational history turns by generating context and question entities, pruning irrelevant turns based on similarity, and re-ranking the remaining turns via an attention-based mechanism to ensure relevance to the current question. Furthermore, a dynamic history selection approach was utilized in \cite{2024_04} to retain only the most relevant conversational turns based on shared entities with the current question. A BART-based sequence-to-sequence model generates question and context entities, and selects historical turns with overlapping entities, thereby reducing noise and improving retrieval quality. A dynamic history selection approach was also applied in \cite{RAG3}, wherein the conversational history was adaptively adjusted at each conversational turn to determine whether external knowledge was needed. In this approach, the conversational context was constructed using all the conversational turns up to the current user turn. A gating function was then applied to this context to decide if the retrieved external knowledge should be included. This decision was made by the RAGate model, which learned to selectively incorporate external knowledge based on the evolving conversational context.

\subsection{Question Understanding}

Question understanding is a fundamental aspect of ConvQA systems. It encompasses various steps aimed at comprehending and interpreting user queries to provide accurate and relevant responses. Unlike traditional single-turn question answering, ConvQA must resolve complex challenges, i.e., coreference resolution, ellipsis, and ambiguity, which are common in multi-turn interactions \cite{CQAHistory1}. As a result, understanding the user’s intent often requires analyzing not only the current question but also the broader conversational context. This step is essential to ensure continuity, maintain coherence, and enable the ConvQA systems to produce accurate and contextually relevant answers. In ConvQA, the popular question understanding approaches can be categorized as mentioned below.

\subsubsection{Question Rewriting}

\begin{figure}[!t]
    \centering
    \includegraphics[width=1.01\textwidth]{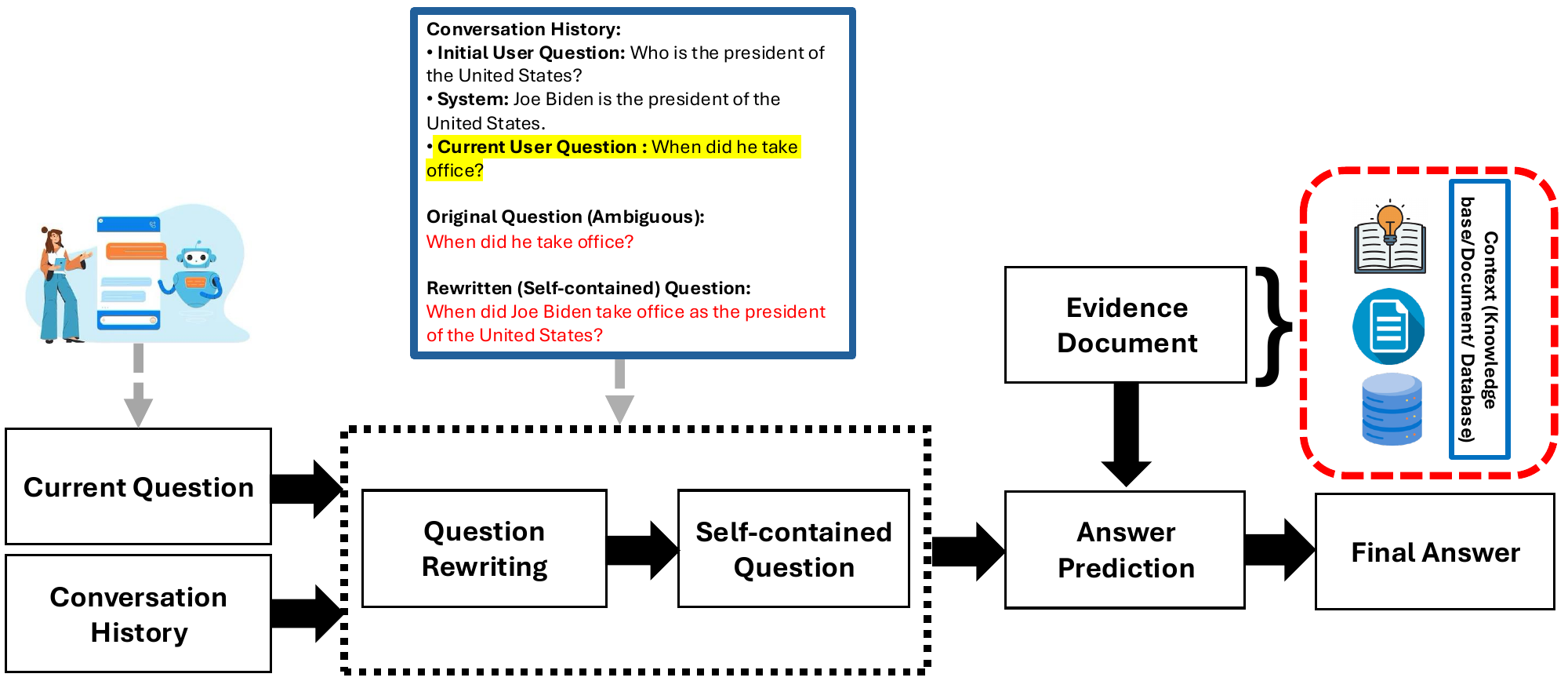} % Adjust the width as needed
     \vspace{-1pt} % Reduce the space between the image and caption
   \captionsetup{justification=centering}
    \caption{Architecture of a conversational question answering system with a question rewriting module.}
    \label{fig:qr_model}
\end{figure}

Question rewriting specifically focuses on rephrasing the original user question to make it clearer and more understandable for the ConvQA systems. Question rewriting often involves simplifying the language, correcting grammatical errors, or changing the structure of the question while preserving its original intent \cite{QRIntro,EntireHistory1, KTurn1}. Figure~\ref{fig:qr_model} delineates a question rewriting model, where the current question is rephrased using the conversational history to create a self-contained question that incorporates all necessary context. This rewritten question enables retrieval of relevant answer phrase from an evidence document or a context paragraph, thereby leading to accurate answer prediction without relying on historical conversational turns. The goal is to produce a more precise and unambiguous question that the ConvQA system can handle more effectively. In the traditional ConvQA setting, most research used the question rewriting technique. This conventional technique was necessary as earlier LLMs, i.e., BERT-based LLMs could not understand ambiguous questions accurately. However even in the past, some authors \cite{Zaib2023KeepingAnswering,ZaibLearningtoselect} argued that
while question rewriting can enhance the question’s clarity, it often detaches the question from
its conversation context by making the question context-independent. This potentially leads to a loss of valuable information. A ConvQA system architecture was presented in \cite{KTurn1} that consisted of separate question rewriting and QA models. The proposed question rewriting model significantly improved the QA performance on the Text REtrieval Conference Conversational Assistance Track (TREC CAsT) and 
Question Answering in Context (QuAC) datasets, thereby making questions explicit and context-independent. A ConvQA system was envisaged in \cite{EntireHistory1} for the Search-Oriented Conversational AI (SCAI) shared task by focusing on evaluating various question rewriting approaches. The authors showed that question rewriting can make conversation context-independent and improves the retrieval and generation of accurate answers. Additionally, question rewriting was further utilized in \cite{Rashid2024NORMY:Answering} to convert the final user question into a self-contained question using a neural coreference resolution model. This ensures that coreference and ellipses are resolved, thereby making the question independent of previous conversational turns. Unlike other modules, the reader module benefits from a narrower and well-formed query rather than full conversational history, thereby improving answer extraction accuracy.

\subsubsection{Question Reformulation}

Question reformulation is another question understanding approach used in ConvQA systems. This approach involves rephrasing or altering the original user question to enhance its alignment with the information available in the ConvQA system's knowledge base or database, thereby improving the accuracy of the retrieved answers. Effective question reformulation can significantly enhance the precision and relevance of the answers retrieved. A new framework entitled, Reign, was introduced in \cite{Robust} which used question reformulation to improve the robustness of the ConvQA system. Question reformulations were selected using deep reinforcement learning to enhance QA performance by training on diverse question variations. The framework introduced a reinforcement learning-based Reformulation Category Selector (RCS) and a BART-based Reformulation Generator (RG) to create intent-preserving variants. These reformulations were selectively incorporated into training, thereby leading to improved robustness against surface-level question changes and enhanced generalization to unseen benchmarks. Furthermore, question reformulation was employed in \cite{RAG1Correct} as a question understanding technique to resolve ambiguities arising from context-dependent references in conversational queries. The authors introduced a Conversational Question Refiner module within their Conversation-level Retrieval-Augmented Generation (ConvRAG) framework, which utilizes a fine-tuned LLM to transform implicit questions into explicit and self-contained versions. This reformulated question enhances the ConvQA system’s ability to accurately retrieve relevant information and generate contextually appropriate responses. Moreover, question reformulation approach was utilized in \cite{QF2} to transform unclear conversational questions into explicit forms using LLM-generated or human-written variations. This helped the ConvQA system to resolve linguistic challenges, i.e., anaphora and ellipsis, thereby making the questions more interpretable for the QA model. To further enhance performance, a teacher-student architecture was employed, where the teacher model learned from human-written reformulations and the student mimicked them using LLM-generated reformulations. This distillation-based strategy allowed the model to approximate human-level performance even when relying solely on synthetic reformulations during testing.

\subsubsection{Named Entity Recognition}

Named Entity Recognition (NER) is crucial for identifying and categorizing entities, i.e., names of people, organizations, locations, dates, and other relevant information within conversational turns \cite{hu2024deep}. Recognizing these entities helps to understand the context and specifications of a user question which is essential for providing accurate responses in ConvQA. NER was utilized in \cite{Zaib2023KeepingAnswering} to identify and extract context and question entities from the conversational history. This process enhanced the understanding of the current question by capturing relevant entities that may be implicitly referenced in the conversation. By generating these structured representations of entities, the ConvQA system could better interpret incomplete or ambiguous questions, thereby ensuring accurate and contextually relevant answers without relying on a separate question rewriting model. NER was further utilized in \cite{ZaibLearningtoselect} to enhance the understanding of the current question by generating context and question entities from the entire conversational history. These entities helped the ConvQA system to identify relevant conversational turns by comparing them with the entities in the current question. By pruning conversational turns that do not share similar entities, the ConvQA system ensured that only the most contextually relevant information was retained, thereby improving the accuracy and relevance of the answer prediction. This approach aided in better capturing the essence of the conversation, particularly in scenarios involving incomplete or ambiguous questions. Furthermore, NER was employed in \cite{2024_01} to identify and extract entities from conversational questions, which are then used to construct and update the entity transition graph. These recognized entities serve as inputs for topic entity prediction and query generation. The authors further discussed errors caused by incorrect entity recognition, thereby highlighting the importance of accurate NER in understanding user intent.

\subsubsection{Semantic Parsing}

Semantic parsing involves transforming natural language questions into a structured, machine-understandable format, i.e., logical form or a query language \cite{sym16091201,wu2024large}. This structured representation enables the ConvQA systems to accurately interpret the user's intent, thereby facilitating precise retrieval of relevant information or generating appropriate responses. In ConvQA, semantic parsing plays a critical role in understanding complex and context-dependent questions by capturing the underlying meaning of the question and mapping it to the relevant data sources or knowledge bases, thereby improving the ConvQA system's ability to provide accurate and contextually relevant answers. Semantic parsing was applied in \cite{semantic1} to transform natural language questions into executable SPARQL Protocol and RDF Query Language (SPARQL) queries over a knowledge graph. The authors developed two approaches, i.e., employing dynamic vocabularies derived from knowledge graphs to generate complete SPARQL queries, and predicting SPARQL query templates which were then populated with relevant entities, relations, and types. These two approaches addressed challenges, i.e., large vocabulary handling, conversational context modeling, and query generation involving multiple entities, thereby ultimately enabling accurate question answering over complex knowledge graphs. Semantic parsing was further employed in \cite{semantic2} to evaluate the ability of LLMs to convert natural language questions into SPARQL queries in a ConvQA setting. By interpreting the user's intent and the context within the conversation, semantic parsing allowed the ConvQA system to generate structured queries that directly interact with the knowledge graph, thereby improving the accuracy and relevance of the generated answers. Moreover, semantic parsing was employed in \cite{KG2} to translate each user questions into a logical form that can be executed over a knowledge graph. The framework used a grammar-guided decoder within a transformer-based architecture to generate these logical forms, thereby enabling structured understanding of complex conversational questions.

\subsubsection{Attention Mechanism}

Attention mechanism in ConvQA is a pivotal technique that enables models to focus on the most relevant parts of the input, i.e., specific words or phrases in a question when generating responses or retrieving information \cite{bai-etal-2024-auto}. By dynamically assigning weights to different parts of the input question, the attention mechanism helps the ConvQA system discern which aspects of the conversational history are most important for understanding the user's current question. This technique is especially useful in ConvQA as it allows the system to effectively handle long or complex conversations, thereby ensuring that the responses are more accurate and contextually aligned with the user's needs. For instance, an attention mechanism was utilized in \cite{KTurn2} by integrating both inter-attention and self-attention within a novel model entitled, SDNet, to enhance the question understanding. Inter-attention helped align and extract relevant information between the question and passage while self-attention allowed the model to relate different parts of the question to each other, thereby facilitating coreference resolution and concept carry-over across the conversation. This dual attention approach combined with BERT embeddings, enabled the model to capture and comprehend the intricate context of the conversation more effectively. Moreover, the self-attention mechanism was employed in \cite{Attention2} as a combination of Bidirectional Gated Recurrent Unit (Bi-GRU) and Conditional Random Fields (CRF) to enhance question understanding by focusing on the most relevant parts of the input sequence. Specifically, the attention mechanism allowed the model to focus on the most significant parts of the text, thereby giving higher weights to relevant tokens within the user’s input. This helped to identify user intent and extract important entities efficiently by determining which words in the sequence were more critical. By integrating attention with the Bi-GRU and CRF models, the system effectively managed contextual information, thereby improving question understanding in ConvQA. The attention mechanism was employed in \cite{2024_02} through the integration of SparseHopfield layers, which compute attention weights using Sparsemax activation. This mechanism aligns the user question with stored memory patterns, thereby effectively highlighting relevant contextual information for sub-question generation and answer refinement. By associating queries with the most relevant knowledge segments, it enhances the model’s ability to understand and decompose complex user questions.

\subsection{Answer Prediction}

Answer prediction in ConvQA systems is a critical process where the system generates or retrieves the most appropriate response based on the input question. This component leverages a combination of techniques to enhance the accuracy, relevance, and context-awareness of the answers provided. As ConvQA systems evolve, the integration of advanced models and methodologies in answer prediction continues to push the boundaries of what these systems can achieve, thereby ensuring more effective and meaningful interactions with users. Recent advancements also emphasize adaptive mechanisms that tailor answers based on evolving conversational cues. These developments contribute to a more contextually fluent and user-aligned answering process across diverse domains. The answer prediction process can be realized using various approaches as delineated below.

\subsubsection{Retrieval-based Approach}

The retrieval-based approach in ConvQA systems involves identifying and retrieving the most relevant answer from a predefined set of responses or a large corpus of documents based on the input question. This approach relies heavily on Information Retrieval (IR) techniques where the system first ranks documents or sentences according to their relevance to the user question and then selects the top-rank answer as the final answer. Traditional IR techniques include models, i.e., Term Frequency-Inverse Document Frequency (TF-IDF) and  Best Matching 25 (BM25) \cite{IR}. Retrieval-based systems are particularly effective when the correct answer already exists in the corpus, thereby making them efficient for tasks requiring high precision and accuracy in providing fact-based responses. A retrieval-based approach was implemented in \cite{EntireHistory2} by utilizing question rewriting to convert conversational questions into self-contained questions. These rewritten questions were passed to a QA model which retrieved relevant answer spans directly from provided evidence documents. The model extracted the most appropriate answer based on the rewritten question and the conversational history. Moreover, the retrieval-based approach was utilized in \cite{Zaib2023KeepingAnswering} to predict answers using various BERT-based models that process the given conversational context and question. The ConvQA model generated structured representations for the conversational turns and then retrieved the most relevant answer span from the context based on these structured representations. Additionally, a new approach entitled, Phrase Retrieval for Open-domain Conversational Question Answering (PRO-ConvQA) \cite{JeongPhraseLearning} was proposed as a retriever-based method for open-domain ConvQA that eliminates the conventional retriever-reader pipeline. Instead of retrieving passages and using a reader for answer prediction, their method directly retrieves answer phrases based on dense phrase retrieval. To enhance retrieval accuracy, this research study employed contrastive learning to model conversational dependencies, thereby ensuring better phrase selection aligned with conversational context. Their approach outperforms retriever-reader pipelines by reducing error propagation and improving efficiency.

\subsubsection{Generative-based Approach}

The generative-based approach in ConvQA systems involves using advanced LLMs to generate answers from scratch based on the input query. Unlike the retrieval-based approach that relies on pre-existing data, the generative-based approach generates responses by predicting the next word in a sequence. This approach allows for more flexible and creative answers, especially in scenarios where the exact answer is not in the training data. Generative models, i.e., GPT, Gemini, and T5, are particularly useful in open-domain settings where a wide range of potential questions and answers need to be handled dynamically. For instance, the generative-based approach was deployed in \cite{EntireHistory1} to generate answers by leveraging the Pegasus model which was designed for abstractive summarization. After retrieving relevant passages using the BM25 model, the ConvQA system inputs these passages along with the rewritten question into the Pegasus model. The Pegasus model then generated the answer by creating a summary that encapsulates the key information from the retrieved passages, thereby synthesizing it into a coherent and contextually relevant response. This generative process allowed the ConvQA system to construct new, context-aware answers rather than merely selecting from pre-existing text. Moreover, a generative-based approach was utilized in \cite{generative2} to predict answers by employing a bi-directional encoder and a prefix autoregressive decoder within its system. The model generated responses by first encoding the medical conversational history and then predicting the next sequence of text based on this encoded context. The generative process also integrated knowledge reasoning from medical entities and symptoms to produce accurate and contextually relevant answers, thereby simulating a natural conversational flow between the user and the system. Furthermore, a generative-based approach was envisaged in \cite{2025_2} for answer prediction by prompting LLMs to generate yes/no sub-questions from policies and compose them into logical formulae. These generated formulae are evaluated to derive a final answer.

\subsubsection{Retrieval-augmented Generation Approach}

Retrieval-augmented Generation (RAG) is indeed part of the broader topic of answer retrieval and generation. RAG combines elements of both retrieval and generative approaches to improve the accuracy and relevance of the generated responses \cite{RAG1,filice2025generatingdiverseqabenchmarks}. The primary goal of RAG is to enhance the quality of generated answers by leveraging retrieved documents or passages as a contextual reference. This approach bridges the gap between purely retrieval-based and purely generative-based models, thereby combining both strengths to produce more accurate and contextually appropriate answers. A new approach entitled, Conversation-level RAG (ConvRAG), was introduced in \cite{RAG1Correct} which enhanced answer prediction by integrating fine-grained retrieval to obtain relevant context from external sources and employing a self-check mechanism to filter out irrelevant information. This combination led to more accurate responses by leveraging retrieved information from historical conversational turns, thereby making ConvRAG well-suited for multi-turn conversational settings. A RAG-based approach entitled, RAGate, was further utilized in \cite{RAG3} to dynamically manage the retrieval and integration of external knowledge in responses. By leveraging retrieved knowledge snippets and using models, i.e., RAGate-MHA, the system predicted and generated more accurate and contextually relevant answers, particularly in scenarios where the conversational context alone may not be sufficient to produce a high-quality response. This approach balanced the need for external knowledge with the generation capabilities of a LLM. Furthermore, a RAG-based approach was utilized in \cite{2025_1}, where answer prediction is grounded in retrieved evidence to mitigate hallucinations and enhance factual accuracy. The ConvQA system first reformulates questions to improve retrieval, then selects relevant evidence from the top 500 documents using a fine-tuned LLM. The final answer is generated by conditioning on this filtered evidence, thereby ensuring that responses are both contextually appropriate and factually supported. The entire process is optimized end-to-end using Direct Preference Optimization (DPO) without relying on human-labeled intermediate data.

\begin{sidewaystable}[htbp]
\scriptsize	
\captionsetup{justification=centering}
\caption{Key components of conversational question answering systems (history selection, question understanding, and answer prediction).}
\centering
\begin{tabular}{
>{\centering\arraybackslash}m{2.2cm}
|>{\centering\arraybackslash}m{1cm}
|>{\centering\arraybackslash}m{1.2cm}
|>{\centering\arraybackslash}m{1cm}
|>{\centering\arraybackslash}m{1cm}
|>{\centering\arraybackslash}m{1cm}
|>{\centering\arraybackslash}m{1cm}
|>{\centering\arraybackslash}m{1cm}
|>{\centering\arraybackslash}m{1cm}
|>{\centering\arraybackslash}m{1cm}
|>{\centering\arraybackslash}m{1cm}
|>{\centering\arraybackslash}m{1cm}
|>{\centering\arraybackslash}m{1cm}
|>{\centering\arraybackslash}m{1cm}
}
\hline\hline
\multirow{6}{*}{\small\textbf{References}} 
& \multicolumn{4}{c|}{\small\textbf{History Selection}} 
& \multicolumn{5}{c|}{\small\textbf{Question Understanding}} 
& \multicolumn{4}{c}{\small\textbf{Answer Prediction}} \\ 
\hline

%\normalsize\textbf{References} 
& \rotatebox{90}{\scriptsize\parbox{2.5cm}{\centering \textbf{K-Turn History \\ Selection}}} 
& \rotatebox{90}{\scriptsize\parbox{2.5cm}{\centering \textbf{Immediate Turn \\ History Selection}}} 
& \rotatebox{90}{\scriptsize\parbox{2.5cm}{\centering \textbf{Entire History \\ Selection}}} 
& \rotatebox{90}{\scriptsize\parbox{2.5cm}{\centering \textbf{Dynamic History \\ Selection}}} 
& \rotatebox{90}{\scriptsize\parbox{2.5cm}{\centering \textbf{Question Rewriting}}} 
& \rotatebox{90}{\scriptsize\parbox{2.5cm}{\centering \textbf{Question\\ Reformulation}}} 
& \rotatebox{90}{\scriptsize\parbox{2.5cm}{\centering \textbf{Named Entity\\ Recognition}}} 
& \rotatebox{90}{\scriptsize\parbox{2.5cm}{\centering \textbf{Semantic\\ Parsing}}} 
& \rotatebox{90}{\scriptsize\parbox{2.5cm}{\centering \textbf{Attention\\ Mechanism}}} 
& \rotatebox{90}{\scriptsize\textbf{Retrieval-based}} 
& \rotatebox{90}{\scriptsize\textbf{Generative-based}} 
& \rotatebox{90}{\scriptsize\textbf{RAG-based}} 
& \rotatebox{90}{\scriptsize\parbox{2.5cm}{\centering \textbf{Knowledge\\ Graph-based}}} 
\\ \hline
\scriptsize \cite{MYACM}: (2025) &  &  & \checkmark &  &  &  & \checkmark &  &  &  & \checkmark  &  & \\ \hline  
\scriptsize \cite{2025_1}: (2025) &  &  & \checkmark &  & \checkmark &  &  &  &  & \checkmark & \checkmark  & \checkmark & \checkmark \\ \hline
\scriptsize \cite{2025_2}: (2025) &  & \checkmark & &   &  & \checkmark  &  & \checkmark &  & & \checkmark & &  \\ \hline
\scriptsize \cite{Robust}: (2024) &  & \checkmark &  &  &  & \checkmark & \checkmark  &  &  & \checkmark &  &  & \checkmark \\ \hline
\scriptsize \cite{RAG1Correct}: (2024) &  & &  \checkmark &  &  & \checkmark &    &  &  & \checkmark & \checkmark & \checkmark &  \\ \hline
\scriptsize \cite{semantic2}: (2024) & \checkmark  &  & &  &  &  &  & \checkmark &  &  & \checkmark &  & \checkmark  \\ \hline
\scriptsize \cite{RAG3}: (2024) &  &  &  & \checkmark &  &  & \checkmark &  & \checkmark &  & \checkmark & \checkmark &  \\ \hline
\scriptsize \cite{Attention2}: (2024) &  &  &  & \checkmark &  &  & \checkmark &  & \checkmark & \checkmark & &  &  \\ \hline
\scriptsize \cite{QF2}: (2024) &  &  & \checkmark &  &  & \checkmark &  &  &  & &  &  & \checkmark \\ \hline
\scriptsize \cite{Rashid2024NORMY:Answering}: (2024) &  &  &  & \checkmark & \checkmark &  &  &  &  &\checkmark &  &  & \\ \hline
\scriptsize \cite{2024_01}: (2024) &  &  &  & \checkmark &  &  & \checkmark &  &  & \checkmark &  & & \checkmark \\ \hline
\scriptsize \cite{2024_02}: (2024) &  &  &  & \checkmark &  &  &   &  & \checkmark & \checkmark & \checkmark  & \checkmark  & \\ \hline
\scriptsize \cite{2024_03}: (2024) & \checkmark  &  &  &  & \checkmark &  &   &  &  & \checkmark & \checkmark  & \checkmark  & \\ \hline
\scriptsize \cite{2024_04}: (2024) &  &  &  & \checkmark &  &  & \checkmark  &  &  & \checkmark &   &  & \\ \hline
\scriptsize \cite{Zaib2023KeepingAnswering}: (2023) &  &  &  & \checkmark &  &  & \checkmark & &  & \checkmark &  &  &  \\ \hline
\scriptsize \cite{ZaibLearningtoselect}: (2023) &  &  &  & \checkmark &  &  & \checkmark & &  \checkmark & \checkmark & &  &  \\ \hline
\scriptsize \cite{JeongPhraseLearning}: (2023) &  &  &  & \checkmark &  &  &  &  & \checkmark & \checkmark &  &  &  \\ \hline
\scriptsize \cite{semantic1}: (2023) &  & \checkmark &  &  &  &  &  & \checkmark &  &  &  &  & \checkmark \\ \hline
\scriptsize \cite{EntireHistory1}: (2022) & \checkmark &  &  &  & \checkmark &  &  &  &  & \checkmark & \checkmark &  &  \\ \hline
\scriptsize \cite{EntireHistory2}: (2022) &  &  & \checkmark &  & \checkmark &  &  &  &  & \checkmark &  &  &  \\ \hline
\scriptsize \cite{generative2}: (2022) &  &  &  & \checkmark &  &   & \checkmark  &  &  &  & \checkmark  &  & \\ \hline
\scriptsize \cite{CL12}: (2022) &  &  & \checkmark &  &  &   &   &  &  \checkmark &  & \checkmark  &  & \checkmark \\ \hline
\scriptsize \cite{EXCORD}: (2021) &  &   & \checkmark &  &  \checkmark &  &  &  &  & \checkmark &  &  &  \\ \hline
\scriptsize \cite{Qiu2021ReinforcedAnswering}: (2021) &  &  &  &  \checkmark &  &  &  &  & \checkmark  & \checkmark &  &  &  \\ \hline
\scriptsize \cite{KG2}: (2021) &  & \checkmark  &  &  &  &  & \checkmark  &  \checkmark  & \checkmark  &  &  &  & \checkmark \\ \hline
\scriptsize \cite{KTurn1}: (2021) & \checkmark &  &  &  & \checkmark  &  &  &  &  & \checkmark & \checkmark  &  &  \\ \hline
\scriptsize \cite{KTurn2}: (2019) & \checkmark &  &  &  &  &  &  &  & \checkmark  & \checkmark &  &  &  \\ \hline
\scriptsize \cite{ImmediateTurn1}: (2019) & \checkmark &  &  &  &  &  &  &  & \checkmark  & \checkmark &  &  &  \\ \hline
\scriptsize \cite{ImmediateTurn2}: (2019) & \checkmark &  &  &  &  &  &  &  & \checkmark  & \checkmark &  &  &  \\ 
\hline\hline
\end{tabular}
\label{comparison_table}
\end{sidewaystable}

\subsubsection{Knowledge Graph-based Approach}

The knowledge graph-based approach in ConvQA utilizes structured information stored in a knowledge graph to enhance the accuracy of answer prediction. When a question is posed, the system queries the knowledge graph to retrieve relevant entities and relationships, thereby providing precise and contextually relevant information. This approach allows the ConvQA system to make informed predictions by leveraging the rich, interconnected data in the knowledge graph, which can be particularly useful for answering fact-based or entity-centric questions. By integrating this structured knowledge with other models, this approach can improve the accuracy and relevance of the answers generated in a conversational context. A knowledge graph-based approach was envisaged in \cite{QF2} by employing a reinforcement learning-based model to navigate a knowledge graph and predict answers. The model used question reformulations generated by LLMs to enhance the understanding of each question. Starting from a topic entity in the knowledge graph, the reinforcement learning agent explored the graph, guided by a policy network to identify the correct answer entity based on the context provided by the conversational history and the reformulated questions. This approach ensured accurate multi-turn conversational question answering by leveraging the structured information within the knowledge graph. A knowledge graph-based approach was utilized in \cite{KG2} by introducing a framework entitled, muLti-task semAntic parSing with trAnsformer and Graph atteNtion nEtworks (LASAGNE), for answer prediction by integrating a transformer-based model with Graph Attention Networks (GATs). The transformer-based model generated logical forms while the GATs exploited the relationships between entity types and predicates in the knowledge graph. This combined approach allowed the model to accurately map conversational questions to logical forms that retrieve answers from the knowledge graph, thereby effectively using the structure and connections within the knowledge graph to guide the answer prediction process. Additionally, a novel approach entitled, Path Ranking for Conversational Question Answering (PRALINE), was introduced in \cite{CL12} which was a contrastive learning-based approach for answer prediction in ConvQA over knowledge graphs. Instead of relying on gold logical forms, PRALINE ranks knowledge graph paths using conversational context, including conversational history, fluent responses, and domain information, to identify the most relevant answer entity. The model employed contrastive representation learning to differentiate between correct and incorrect paths, thereby significantly improving knowledge graph-based answer prediction.

Table \ref{comparison_table} visually summarizes how the key components of ConvQA, i.e., history selection, question understanding, and answer prediction, are utilized across the state-of-the-art literature.

\section{Machine Learning Techniques for Conversational Question Answering Systems}

\begin{figure}[!t]
    \centering
    \includegraphics[width=1.0\textwidth]{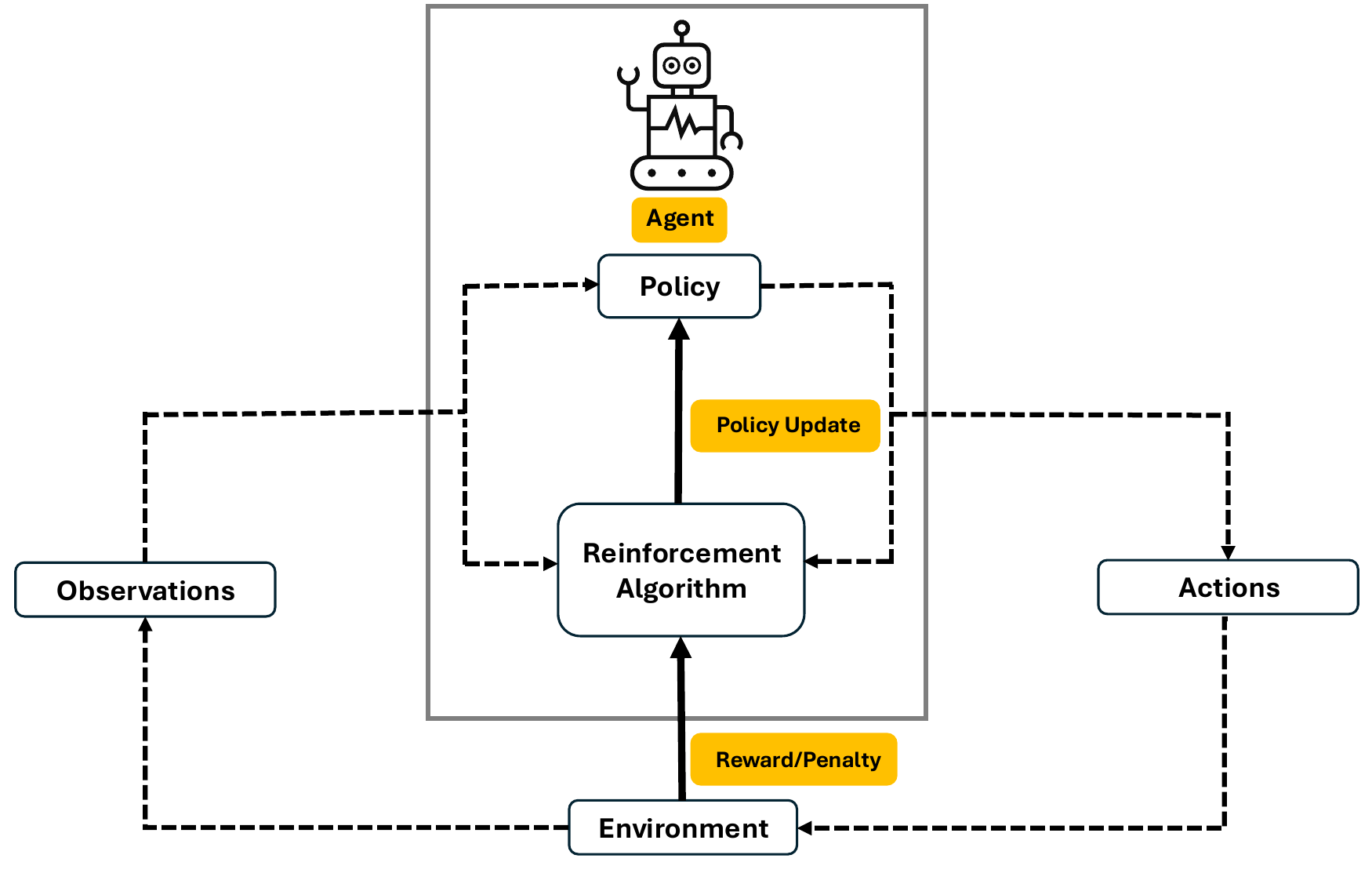} % Adjust the width as needed
   \vspace{10pt} % Reduce the space between the image and caption
    \captionsetup{justification=centering}
    \caption{Reinforcement Learning Framework -- Illustration of an interaction between the agent, environment, and policy through observations, actions, and rewards to optimize learning and decision making.}
    \label{fig:rl_model}
\end{figure}

ConvQA systems have made substantial progress in recent years, driven by the application of advanced machine learning and deep learning techniques. By leveraging machine learning techniques, i.e., Reinforcement Learning (RL), Knowledge Distillation (KD), Contrastive Learning (CL), Active Learning (AL), and Transfer Learning (TL), researchers have developed more sophisticated and effective ConvQA systems. These advancements not only improve answer accuracy and contextual understanding but also enhance efficiency and generalization across domains. This section explores the integration and impact of these techniques on the ConvQA system's performance and capabilities.

\subsection{Reinforcement Learning}

RL integrates feedback mechanisms into the training process of ConvQA systems, thereby allowing the ConvQA systems to learn and adapt from interactions over time. In an RL framework, the feedback is used as reward signals that guide the learning process \cite{khadivi2025deep,RLandLLM,RLLLM2}. The positive and negative feedback helps the ConvQA system to adjust its responses to maximize positive outcomes. This enables the system to continuously improve by incorporating feedback into its decision-making process to give more accurate responses \cite{Chen2022ReinforcedAnswering}. Figure \ref{fig:rl_model} illustrates the general structure of a RL framework, which highlights the interaction between the agent, environment, and policy. The agent observes the environment’s current state and performs actions based on its policy which is a strategy that determines how the agent responds to various situations. Once the action is executed, the environment transitions to a new state and generates a corresponding reward or penalty, which serves as feedback for the agent. This feedback loop enables the RL algorithm to iteratively update the policy, thereby improving the agent’s decision-making process over time. This framework is particularly valuable in scenarios, i.e., ConvQA, where maintaining coherence and relevance across multi-turn conversations is critical. To improve ConvQA system performance, RL was utilized to backtrack and select relevant conversational history turns in \cite{Qiu2021ReinforcedAnswering}. The proposed approach treated the selection of historical conversational turns as a sequential decision-making process with an RL agent. It learned to filter the irrelevant conversational history turns and focus on the most pertinent conversational turns to enhance the accuracy of the answers. RL was employed in \cite{2024_06RL} to handle multi-turn legal conversations over a knowledge graph. An RL agent navigates the knowledge graph based on the current question and conversational history to locate accurate answers, even when input questions are ambiguous or noisy. The model formalizes the problem as a Markov Decision Process (MDP) and applies the RL for training. Furthermore, the RL was employed in \cite{Robust} to enhance the robustness of ConvQA systems. By leveraging RL, the authors developed a strategy to adaptively select training examples for question rewriting that maximize model performance, thereby improving the ConvQA system's ability to handle challenging conversational contexts.

\subsection{Knowledge Distillation}

\begin{figure}[!t]
    \centering
    \includegraphics[width=1.0\textwidth]{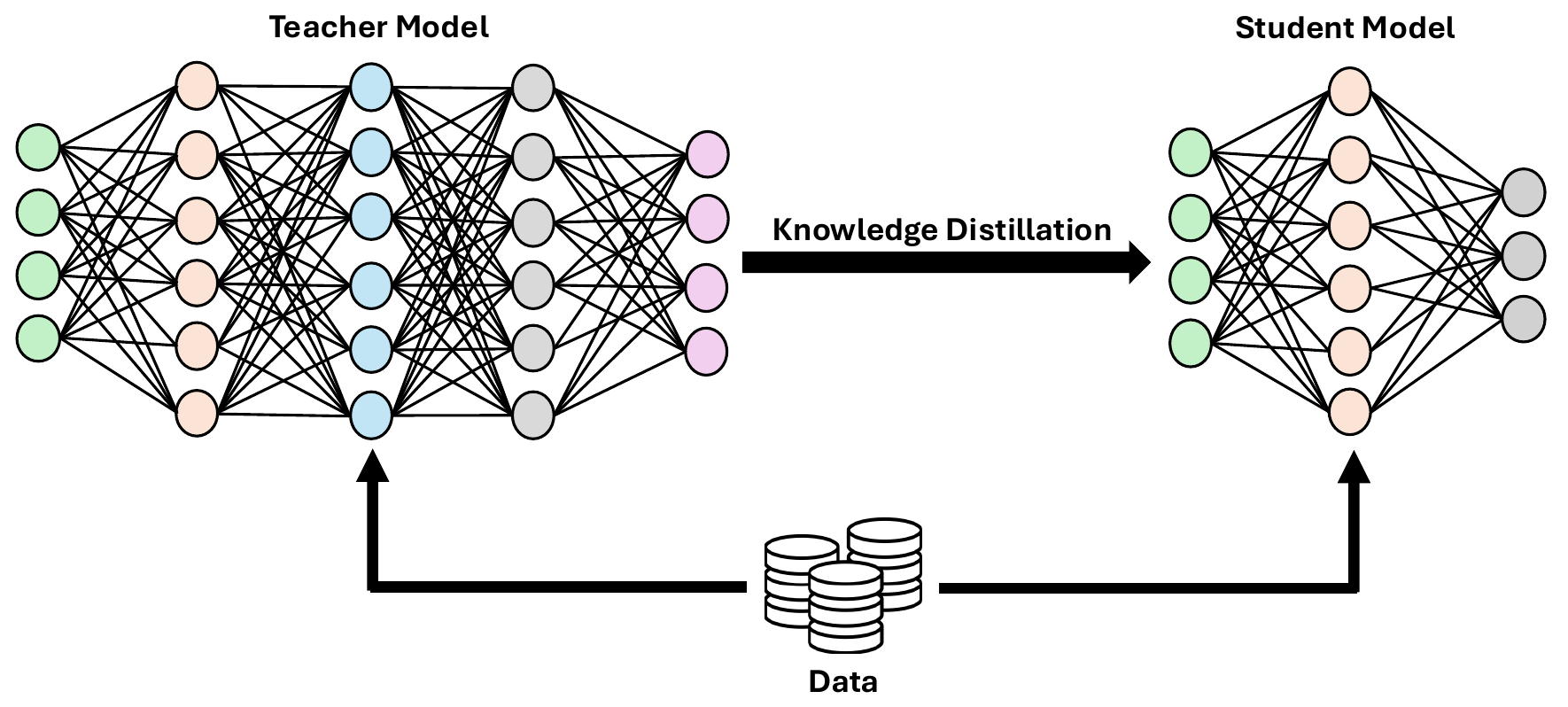} % Adjust the width as needed
   \vspace{-10pt} % Reduce the space between the image and caption
   \captionsetup{justification=centering}
    \caption{Knowledge Distillation Framework -- The teacher model trained on a dataset, transfers its learned knowledge to a simplified student model, thereby enabling efficient deployment while retaining performance.}
    \label{fig:kd_model}
\end{figure}

KD is a technique in machine learning where a large and complex model (teacher) transfers its knowledge to a smaller, more efficient model (student) \cite{SUN2025111095,KDLLM,KD3}. The teacher model provides soft labels or additional insights that help guide the training of the student model, thereby enabling it to achieve high performance with reduced computational resources. Figure \ref{fig:kd_model} illustrates the concept of KD, where a high-capacity teacher model is used to guide the training of a smaller student model. Both models are trained on the same dataset, but the student model mimics the behavior of the teacher model through soft target outputs, thereby enabling it to learn the essential patterns captured by the teacher model. This process helps in creating lightweight models suitable for real-world deployment without significant loss of accuracy. The arrows indicate the flow of knowledge, from the teacher model to the student model, with the data being the input for both. KD was applied in \cite{KD1} to enhance a RoBERTa-based ConvQA model by transferring knowledge from multiple teacher models through averaged soft labels. It was integrated with rationale tagging and adversarial training in a multi-task learning framework. This combination significantly improved the model’s robustness and generalization. The resulting ConvQA system achieved state-of-the-art performance on the CoQA benchmark without using additional training data. Moreover, KD was applied in \cite{KDandAL} to compress large BERT-based teacher models into lightweight student models, i.e., QANet, BiDAF, and Match-LSTM. The student models were trained using both hard labels and soft logits from the teacher to retain performance. An interpolation strategy addressed tokenizer mismatches between teacher and student models. KD significantly improved accuracy, robustness, and inference efficiency, thereby achieving state-of-the-art performance with only a fraction of the original model’s parameters. Furthermore, KD was utilized in \cite{KD2} to train lightweight student models, i.e., BiLSTM and TextCNN, using the PAL-BERT model as the teacher model. The distillation combined soft-label alignment, representation matching, and attention-based loss to transfer knowledge effectively. This approach significantly reduced inference time while maintaining strong answer prediction performance.

\subsection{Contrastive Learning}

\begin{figure}[!t]
    \centering
    \includegraphics[width=0.85\textwidth]{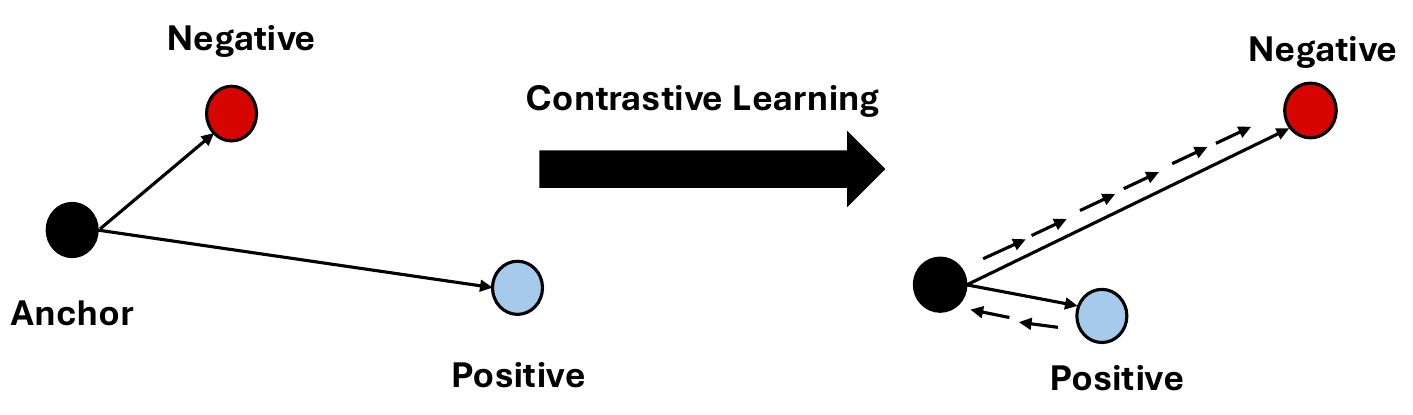} % Adjust the width as needed
   \vspace{10pt} % Reduce the space between the image and caption
    \captionsetup{justification=centering}
    \caption{Contrastive Learning Framework -- The model learns representations by bringing the anchor and positive sample closer together in the latent space while pushing the negative sample further away.}
    \label{fig:cl_model}
\end{figure}

CL is a technique used in machine learning to learn representations by comparing data points. The main purpose of CL is to bring similar instances closer together in the embedding space and push dissimilar instances further apart \cite{CL3}. This is achieved by creating positive (similar) and negative (dissimilar) pairs of data \cite{XU2025102500,gao2024customizing}. The model was then trained to minimize the distance between positive pairs while maximizing the distance between negative pairs. In the context of ConvQA, this can be used to improve the quality of answers by ensuring that similar questions yield similar answers, thereby enhancing the consistency and accuracy of the system. Figure \ref{fig:cl_model} demonstrates the core principle of CL, which is to create a latent space where similar data points (anchor and positive) are closely aligned, and dissimilar ones (anchor and negative) are distinctly separated. In the initial phase, the anchor, positive, and negative samples are positioned arbitrarily in the space. Through the learning process, the model adjusts the representation such that the positive sample (e.g., a relevant response or similar item) becomes closer to the anchor, while the negative sample (e.g., an irrelevant response or dissimilar item) moves farther away. This process enables the model to learn robust and discriminative embeddings that can generalize effectively to downstream tasks. CL was employed in \cite{JeongPhraseLearning} to model the dependencies between consecutive conversational turns in a conversation. This approach ensured that the representations of the current and previous conversational contexts were similar. This helped to retrieve relevant phrases for the current context by maximizing the similarities between consecutive conversational turns and minimizing the similarities with irrelevant contexts. This strategy enhanced the ConvQA system's ability to reflect the previous conversational turns when retrieving phrases, thereby improving the overall performance in ConvQA tasks. Furthermore, CL was utilized in \cite{CL12} to rank knowledge graph paths by learning representations that distinguish between correct (positive) and incorrect (negative) paths. This approach incorporated the entire conversational history and domain information to jointly learn representations of the conversation and knowledge graph paths. By contrasting these representations, the model effectively ranked the knowledge graph paths, thereby significantly improving key metrics, i.e., Mean Reciprocal Rank (MRR) and Hit@5. Moreover, CL was employed in \cite{2024_04} to improve dense passage retrieval in ConvQA by minimizing the distance between relevant passages and maximizing it for irrelevant ones. This enabled more accurate retrieval of contextually aligned passages using curated conversational history. The approach significantly boosted answer accuracy by reducing retrieval noise and enhancing passage relevance within multi-turn conversations.

\subsection{Active Learning}

\begin{figure}[!t]
    \centering
    \includegraphics[width=1\textwidth]{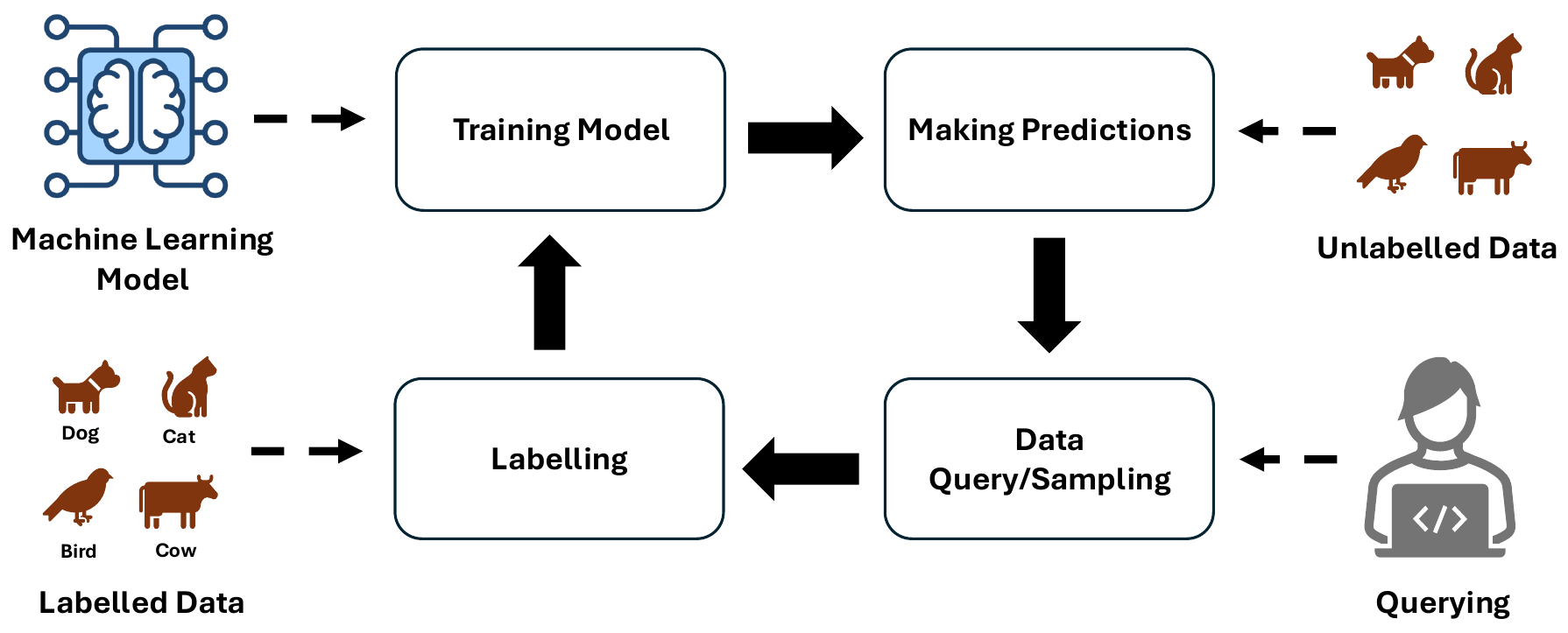} % Adjust the width as needed
   \vspace{10pt} % Reduce the space between the image and caption
    \captionsetup{justification=centering}
    \caption{Active Learning Framework -- The process iteratively trains a model using labeled data, queries informative samples from unlabeled data for labeling, and refines the model through feedback to improve prediction performance.}
    \label{fig:al_model}
\end{figure}

AL is an approach used in machine learning, particularly within NLP to improve the accuracy of models while reducing the need for extensive labeled training data. The key idea behind AL is that a model can actively select the most informative data points from an unlabeled pool to be annotated, thereby minimizing the number of required annotations while maximizing model performance \cite{ALSurvey}. Figure \ref{fig:al_model} represents the AL cycle, which is commonly used to maximize the efficiency of machine learning models when labeled data is scarce. It begins with an initial machine learning model trained on available labeled data. This model makes predictions on a pool of unlabeled data. A querying strategy then selects the most informative samples for manual labeling (e.g., by a human annotator). These newly labeled samples are added to the labeled dataset to retrain and refine the model, thus improving its predictive accuracy iteratively. This loop continues until the model achieves satisfactory performance or the labeling budget is exhausted. The diagram effectively highlights the interaction between labeled data, unlabeled data, and the human annotator in this framework. AL was applied in \cite{KDandAL} to minimize the amount of labeled data required for training a QA model while maintaining high accuracy. This research study employed several AL strategies, i.e., least confidence sampling, margin sampling, entropy-based sampling, and a clustering method to select the most informative unlabeled samples for annotation. This approach allowed the model to achieve competitive performance using only a fraction of the training data compared to traditional methods, thereby reducing annotation costs and improving training efficiency. Moreover, AL was employed in \cite{ActiveLearning2} to improve the performance of a multi-label classifier tasked with detecting ellipsis and coreference in ConvQA datasets. The process involved iteratively training and evaluating the model, and selecting the most uncertain predictions for manual annotation in subsequent training rounds. This method allowed for targeted improvements in model accuracy, particularly in scenarios with limited labeled data by focusing human annotation efforts where those efforts would be most beneficial. Furthermore, AL was employed in \cite{2025_05AL} to identify the most informative conversation samples from LLM-generated conversations for annotation, thereby enabling efficient dataset construction. The AL4RAG framework introduced a retrieval-augmented similarity metric tailored to the RAG setting, thereby improving sample selection quality.

\begin{sidewaystable}[htbp]
\centering
\scriptsize
\caption{Machine learning techniques, references, and summaries.}
\label{tab:ml_summary_table_part1}

\renewcommand{\arraystretch}{1.6}
\setlength{\tabcolsep}{4pt}

\begin{tabular}{
>{\centering\arraybackslash}m{3.5cm}
|>{\centering\arraybackslash}m{1cm}
|m{18cm}
}
\hline\hline
\textbf{\centering Machine Learning Technique} 
& \textbf{\centering Ref.} 
& \multicolumn{1}{c}{\textbf{Summary}} \\
\hline

\multirow{7}{=}{\centering Reinforcement Learning (RL)}
 & \cite{2024_06RL} & RL was utilized to train an agent that navigates knowledge graphs using conversational history and the current question to find accurate answers, even with noisy or unclear questions. \\
\cmidrule{2-3}
 & \cite{Chen2022ReinforcedAnswering} & RL was used to train a question rewriting model that learns from QA system feedback, thereby optimizing rewrites based on rewards, i.e., F1 score and model confidence. \\
\cmidrule{2-3}
 & \cite{Qiu2021ReinforcedAnswering} & RL was utilized in this research to train an agent that selectively backtracks through the conversational history to identify the most relevant question-answer pairs for answering the current question. \\
\cmidrule{2-3}
 & \cite{Robust} & RL was utilized in this research to train a reformulation selector that selects the best question variants using Deep Q-Networks, guided by QA performance as rewards. This enhanced the robustness of the ConvQA system.\\
\hline

\multirow{8}{=}{\centering Knowledge Distillation (KD)}
 & \cite{KD1} & KD was used in this research to train a student model by leveraging soft probability outputs from multiple teacher models. These teacher model outputs served as additional supervision, thereby helping the student model learn more effectively and improve answer prediction in ConvQA. \\
\cmidrule{2-3}
 & \cite{KD2} & KD was utilized in this research to train lightweight student models (BiLSTM and TextCNN) using PAL-BERT as the teacher model. By applying soft-label, representation, and attention-based distillation during fine-tuning, the distilled models significantly reduced inference time while maintaining competitive QA performance. \\
\cmidrule{2-3}
 & \cite{KD3} & KD was utilized by training a teacher model on clean speech and text, which then guided a student model trained solely on noisy Automatic Speech Recognition (ASR) transcriptions. This approach enabled the student to learn robust cross-modal representations, improving answer accuracy in spoken conversational QA settings. \\
\cmidrule{2-3}
 & \cite{KDandAL} & KD was utilized in this research to transfer knowledge from large pre-trained teacher models to smaller student models. This improved ConvQA efficiency by maintaining high accuracy while reducing model size and computational cost. \\
\hline

\multirow{8}{=}{\centering Contrastive Learning (CL)}
 & \cite{JeongPhraseLearning} & CL was employed to enhance phrase retrieval by modeling dependencies between consecutive conversational turns and to rank knowledge graph paths by distinguishing between correct and incorrect representations, thereby improving answer prediction in ConvQA systems. \\
\cmidrule{2-3}
 & \cite{CL12} & CL was utilized in a novel approach to jointly embed conversational context and knowledge graph paths, thereby distinguishing positive and negative paths through contrastive loss. This enabled effective knowledge graph path ranking without relying on gold logical forms. \\
\cmidrule{2-3}
 & \cite{CL3} & CL was utilized to enhance long-context QA by maximizing similarity between questions and their supporting evidence sentences while minimizing it with distractors. This supervised contrastive objective improved both evidence retrieval and answer accuracy across multiple transformer models.\\
\cmidrule{2-3}
 & \cite{2024_04} & CL was employed to enhance dense passage retrieval by pulling relevant passages closer to the user question and pushing irrelevant ones apart in the embedding space. This approach improved passage selection accuracy in open-domain ConvQA. \\
\hline\hline

\end{tabular}
\end{sidewaystable}

\addtocounter{table}{-1} 
\begin{sidewaystable}[htbp]
 % Slightly taller rows

\centering
\scriptsize
\caption{\textbf{Continued}: Machine learning techniques, references, and summaries.}
\label{tab:ml_summary_table_part2}

\renewcommand{\arraystretch}{1.6}
\setlength{\tabcolsep}{4pt}

\begin{tabular}{
>{\centering\arraybackslash}m{3.5cm}
|>{\centering\arraybackslash}m{1cm}
|m{18cm}
}
\hline\hline
\textbf{\centering Machine Learning Technique} 
& \textbf{\centering Ref.} 
& \multicolumn{1}{c}{\textbf{Summary}} \\
\hline

\multirow{6}{=}{\centering Active Learning \\ (AL)}
 & \cite{KDandAL} & AL was utilized to improve ConvQA performance by selecting the most informative unlabeled samples for annotation using strategies, i.e., entropy-based and margin sampling. This approach reduced the labeling effort while maintaining high model accuracy. \\
\cmidrule{2-3}
 & \cite{ActiveLearning2} & AL was employed to iteratively select the most uncertain conversation instances from the CANARD dataset for manual annotation of ellipsis and coreference labels. This strategy enhanced classifier performance by efficiently expanding the labeled training set in low-resource ConvQA settings. \\
\cmidrule{2-3}
 & \cite{2025_05AL} & AL was employed to identify the most informative conversation samples from LLM-generated conversations for annotation, thereby enabling efficient dataset construction. The AL4RAG framework introduced a retrieval-augmented similarity metric to enhance sample selection in the RAG setting. \\
\hline

\multirow{8}{=}{\centering Transfer Learning \\ (TL)}
 & \cite{TL1} & TL was applied by training on English QuAC data and fine-tuning on a Basque ConvQA dataset, thereby enabling effective cross-lingual performance in low-resource settings. \\
\cmidrule{2-3}
 & \cite{TL2} & TL was utilized by fine-tuning pre-trained BERT transformer models from Hugging Face on the SQuAD dataset to perform extractive QA. This approach leveraged existing language understanding capabilities to reduce computational cost and training time. \\
\cmidrule{2-3}
 & \cite{MYACM} & TL was employed in this research by fine-tuning pre-trained LLMs, e.g., RoBERTa, BERT-Large, LLaMA-2 on the coqa chat dataset. This adaptation enabled the models to better understand conversational context and improve answer relevance in ConvQA tasks. \\
\cmidrule{2-3}
 & \cite{Zaib2023KeepingAnswering} & TL was utilized in this research by fine-tuning pre-trained LLMs, i.e., GPT-2 and BART for question rewriting and structured representation generation, respectively. These models were adapted to handle conversational context in ConvQA tasks using the CANARD and QuAC datasets. \\
\hline\hline

\end{tabular}
\end{sidewaystable}

\subsection{Transfer Learning}

\begin{figure}[!t]
    \centering
    \includegraphics[width=0.7\textwidth]{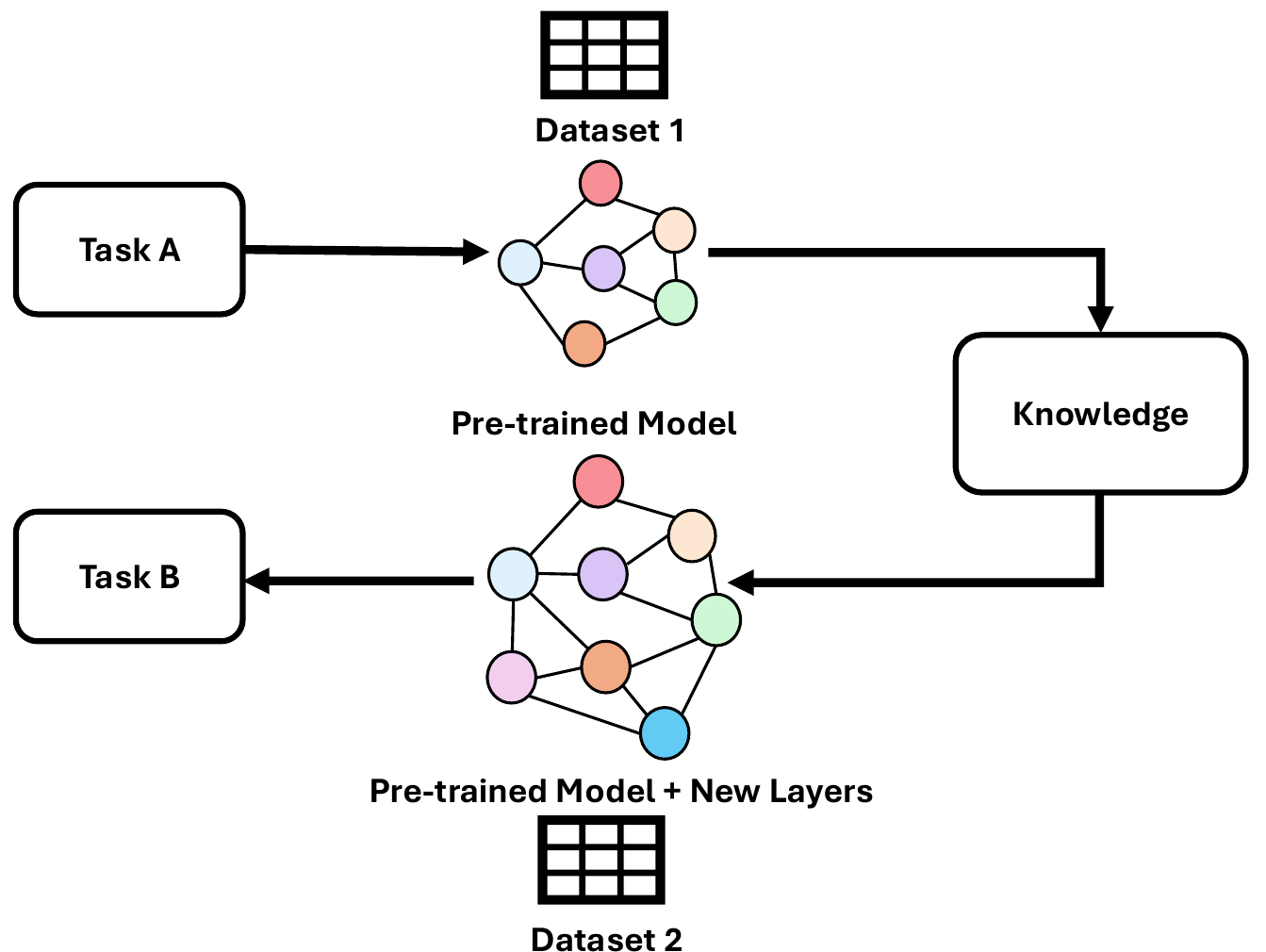} % Adjust the width as needed
   \vspace{10pt} % Reduce the space between the image and caption
    \captionsetup{justification=centering}
    \caption{Transfer Learning Framework -- A pre-trained model, trained on Task A with Dataset 1, leverages its learned knowledge to adapt to Task B with Dataset 2 by fine-tuning additional layers.}
    \label{fig:tl_model}
\end{figure}

TL is a machine learning technique where a pre-trained model developed for a specific task is reused as the starting point for a new but related task \cite{ZHENG2025110562,TLLLM1}. This method leverages the knowledge gained from the initial task, i.e., recognizing common patterns in images or language to improve performance on the new task with less training data. It is particularly useful in fields, i.e., computer vision and NLP where acquiring labeled data is challenging and computationally intensive. TL thus saves time and resources while enhancing model accuracy and efficiency. In the context of ConvQA, TL is particularly useful as it allows models to utilize pre-existing language understanding capabilities to address new challenges without the need for extensive domain-specific training. This approach increases the speed of the development process and enhances the model's ability to handle diverse and complex queries by building on the generalized knowledge from pre-trained models. Figure \ref{fig:tl_model} represents the process of TL, where a model pre-trained on one task (Task A) using a specific dataset (Dataset 1) captures generalizable knowledge. This knowledge is then transferred and reused for a new task (Task B) with a different dataset (Dataset 2). The transfer involves adapting the pre-trained model by fine-tuning it with additional layers tailored to the new task. This framework allows for efficient learning, thereby reducing the need for large amounts of labeled data for the new task, and speeds up convergence by utilizing the pre-trained model's existing features. TL was utilized in \cite{TL1} by applying cross-lingual transfer from English to Basque using pre-trained multilingual BERT models. The authors fine-tuned these models on a small Basque dataset after initial training on the larger English QuAC dataset which allowed them to achieve performance levels comparable to English despite the low-resource Basque scenario. Taking this approach further, TL was applied in \cite{TL2} by leveraging pre-trained BERT transformer models, specifically using Hugging Face’s pre-trained models. These models were fine-tuned on the SQuAD dataset for the specific task of extractive question answering. TL allowed the model to apply knowledge gained from large datasets to the ConvQA task, thereby reducing the computational resources required and avoiding the need to train a model from scratch. Moreover, TL was utilized in \cite{MYACM} to adapt pre-trained LLMs, i.e., RoBERTa, BERT-Large, mDeBERTa-v3, Llama-2-7b, Mistral-7B, and Phi-2, to the task of ConvQA. These LLMs, originally trained on general-purpose corpora, were fine-tuned on the coqa chat dataset to evaluate their proposed Adaptive Context Management (ACM) framework. This TL approach enabled the models to leverage prior linguistic knowledge and improve their performance on multi-turn conversational tasks. 

Table \ref{tab:ml_summary_table_part1} provides a summarized view of previously discussed studies by organizing them according to the machine learning techniques applied in ConvQA.

%%%%%%

\section{Large Language Models for Conversational Question Answering Systems}

\begin{figure}[!t]
    \centering
    \includegraphics[width=1\textwidth]{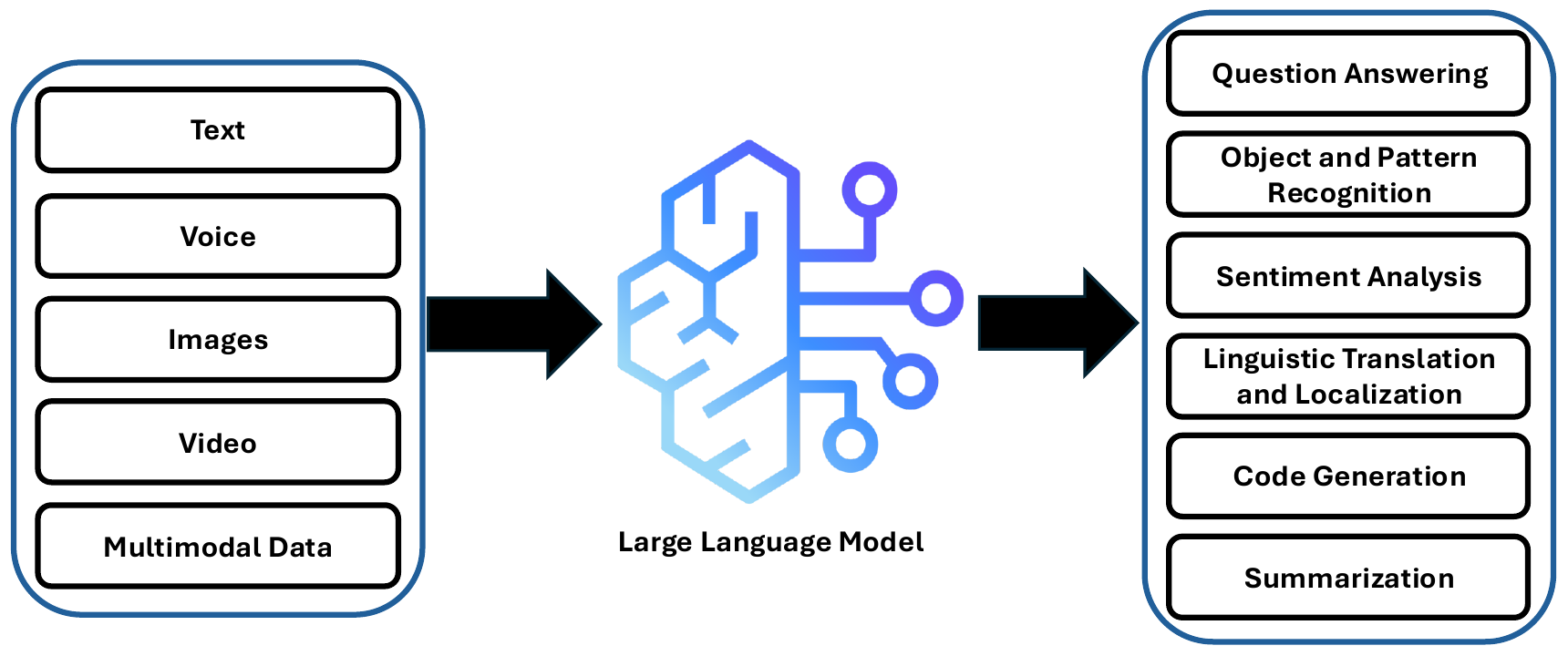} % Adjust the width as needed
   \vspace{9pt} % Reduce the space between the image and caption
    \captionsetup{justification=centering}
    \caption{Comprehensive input-output framework of a large language model.}
    \label{fig:llm_model}
\end{figure}

LLMs that have been trained on extensive datasets have demonstrated a significant impact on NLP \cite{2025_03}. These models can leverage their acquired knowledge across different applications regardless of the particular context in which they were initially trained \cite{xi2025rise,zheng2025towards,li2024survey,2025_04}. These models serve as a foundation for various tasks, i.e., language translation, sentiment analysis, text summarization, QA, and ConvQA. The key LLMs including RoBERTa, GPT-4, Gemini 2.0 Flash, LLaMA 3, and Mistral 7B, known for their vast training on text datasets can be adapted for precise tasks, i.e., QA and ConvQA \cite{Min2023RecentSurvey}. LLMs have shown an amazing ability to combine intricate linguistic and factual knowledge into their parameters as they expand to comprise billions of parameters \cite{LLMEx1,LLMex2}. Figure \ref{fig:llm_model} illustrates the input-output architecture of LLMs by showcasing their ability to process diverse data types and generate intelligent outputs. It highlights five key input modalities i.e., voice, text, images, video, and multimodal data, that LLMs can effectively analyze. The central LLM component serves as the core processing unit, transforming these inputs into meaningful outputs. The outputs include tasks i.e., question answering, object and pattern recognition, sentiment analysis, linguistic translation and localization, code generation, and summarization. This framework demonstrates the versatility and adaptability of LLMs across various domains and applications.

\subsection{Robustly Optimized BERT Approach}
%\vspace{-1ex}

Robustly Optimized BERT Approach (RoBERTa) is an improved version of BERT \cite{RobertaOri}. It was developed by Facebook Inc. RoBERTa addresses the key limitations identified in the original BERT model by optimizing the pretraining process. These optimizations entail extending the training period and employing larger batch sizes, along with utilizing more extensive datasets. A novel framework entitled, Explicit guidance on how to resolve Conversational Dependency (ExCoRD), was proposed in \cite{EXCORD} to resolve conversational dependencies, i.e., anaphora and ellipsis. The framework's effectiveness was showcased through various language model benchmarks. Notably, RoBERTa exhibited a substantial enhancement in performance by achieving an impressive F1 score of 67.7 on the QuAC dataset. The RoBERTa model, as demonstrated in \cite{Robera1}, achieved state-of-the-art results across multiple benchmarks by removing the Next Sentence Prediction (NSP) objective, increasing training duration with larger batches, utilizing dynamic masking, and incorporating a vast new dataset (CC-NEWS). The result demonstrated the RoBERTa model's robustness and effectiveness in natural language understanding tasks.

\subsection{Generative Pre-trained Transformer-4}

OpenAI's Generative Pre-trained Transformer-4 (GPT-4) is a cutting-edge LLM that generates texts with more precision, nuance, and proficiency than its predecessors. GPT-4 handles 25,000 words of text, advanced reasoning, and complex instructions more effectively than GPT-3.5 \cite{gpt4technicalreport}. One of its significant improvements is its ability to follow complex instructions and respond coherently and contextually \cite{gpt4}. The performance of GPT-4 in ConvQA tasks was assessed in \cite{rangapur2024battlellmscomparativestudy} wherein the authors emphasized the substantial enhancements in GPT-4's capacity to produce pertinent, precise, and coherent replies in comparison to other models, i.e., ChatGPT-3, Gemini, Mixtral, and Claude. GPT-4 exhibited exceptional performance in ConvQA applications as evidenced by its rigorous evaluation using metrics, i.e., BLEU and ROUGE. This makes GPT-4 a highly promising candidate. Moreover, GPT-4 was leveraged in \cite{GPT-42} within a Generate-then-Retrieve (GR) pipeline to improve conversational passage retrieval. GPT-4 was employed as a zero-shot learner to generate initial responses based on user queries and conversational context. Additionally, it generated multiple search queries from these responses which were then used to retrieve relevant passages, thereby significantly enhancing retrieval effectiveness for complex queries.

\subsection{Gemini 2.0 Flash}

Gemini 2.0 Flash, a multimodal transformer developed by Google DeepMind, extends conversational AI capabilities by supporting text, code, image, and audio inputs. Unlike traditional LLMs that primarily process text, Gemini 2.0 Flash is designed to understand and reason across multiple modalities simultaneously, thereby enabling richer and more interactive user experiences. This design allows it to interpret complex inputs, i.e., analyzing visual data alongside textual instructions or combining audio signals with written queries. Its architecture supports unified tokenization and cross-modal attention, thereby ensuring contextual coherence when handling diverse input types in real time. In a recent comparative research study \cite{GeminiFlash2.0_2}, Gemini 2.0 Flash demonstrated high performance in multi-turn QA, scientific reasoning, and multilingual conversation tasks. Its "Flash Thinking" mechanism enhances real-time response generation, thereby making it suitable for responsive and context-aware conversational systems. A conversational diagnostic system was introduced in \cite{gemini2.0Flash} leveraging Gemini 2.0 Flash. This research study envisaged the Articulate Medical Intelligence Explorer (AMIE) system that employs a state-aware dialogue framework, thereby dynamically adjusting conversation flow based on intermediate model outputs reflecting patient states and evolving diagnoses. Evaluated against primary care physicians in 105 simulated consultations, AMIE demonstrated superior performance in diagnostic accuracy, history-taking, and empathetic communication across multiple clinical scenarios.

\subsection{Mistral 7B}

Mistral 7B is a highly efficient 7-billion-parameter language model that performs very effectively in reasoning, mathematics, and code generation. It employed innovative techniques, i.e., Grouped Query Attention (GQA) for faster inference and Sliding Window Attention (SWA) to handle longer sequences effectively. Mistral 7B has been optimized to operate in real-time applications, thereby minimizing computational cost while maintaining high performance. Additionally, a fine-tuned version, Mistral 7B – Instruct, excels in instruction-following tasks, outperforming LLaMA 2 13B – Chat model on both human and automated benchmarks \cite{Mistral7BO}. The Mistral-7B model was integrated with graph-based reasoning to enhance ConvQA in \cite{Mistral7B1}. The dynamic graphs were generated by the authors from the retrieved evidence and inserted into the Mistral-7B model, thereby bypassing the token embedding layers. This approach paired with a memory module that tracks and updates past evidence, showed significant reasoning improvements across multiple information sources as validated on the ConvMix dataset. The evaluation of the multilingual capabilities of the Mistral 7B model was conducted in \cite{Mistral7B2} using the MULTIQ benchmark. This benchmark consisted of 27,400 questions in 137 different languages. 
The study assessed the model's language fidelity, ability to respond in the prompted language, and QA accuracy. Notably, Mistral 7B demonstrated impressive language accuracy by responding 62.3\% answers correctly for test cases in the given language.

\begin{sidewaystable}[htbp]
\scriptsize % Smaller text size to fit better
\caption{Comparison of large language models for conversational question answering systems.}
\centering
\renewcommand{\arraystretch}{1.8} % Adjust row height for better vertical centering
\setlength{\tabcolsep}{1.5pt} % Slightly reduce column spacing

\begin{tabular}{
>{\centering\arraybackslash}m{3cm} |
>{\centering\arraybackslash}m{2.6cm} |
>{\centering\arraybackslash}m{2.5cm} |
>{\centering\arraybackslash}m{2.5cm} |
>{\centering\arraybackslash}m{1.2cm} |
>{\centering\arraybackslash}m{1.6cm} |
>{\centering\arraybackslash}m{1.8cm} |
>{\centering\arraybackslash}m{3.8cm} |
>{\centering\arraybackslash}m{2cm}
}
\hline\hline
\textbf{Model} & \textbf{Parameters} & \textbf{Training Tokens} & \textbf{Token Size (Context Length)} & \textbf{Open Source} & \textbf{Base Model} & \textbf{Released Year} & \textbf{Training Dataset} & \textbf{Developer} \\ 
\hline
RoBERTa \protect\cite{RobertaOri} & 355M & 2.2T & 512 & Yes & BERT & 2019 & BooksCorpus, English Wikipedia, CC-NEWS, STORIES, Reddit & Meta (Facebook AI) \\ \hline
GPT-4 \protect\cite{gpt4technicalreport} & 1.76T & 13T & 8K (Standard), 32K (Extended) & No & GPT-3 & 2023 & Common Crawl, WebText, Wikipedia & OpenAI \\ \hline
Gemini 2.0 Flash \protect\cite{gemini2.0Flash} & Not publicly disclosed & Not publicly disclosed & 1M & No & Gemini 1.5 Flash & 2025 & Web documents, books, code, image/audio/video & Google DeepMind \\ \hline
Gemini 2.0 Pro \protect\cite{google2025gemini} & Not publicly disclosed & Not publicly disclosed & 1M (standard), 2M (extended) & No & Gemini 1.5 Pro & 2025 & Web documents, books, code, image/audio/video & Google DeepMind \\ \hline
DeepSeek-R1 \protect\cite{deepseek} & 671B & 14.8T & 128K & Yes & DeepSeek-V3 & 2025 & Instruction datasets, reasoning benchmarks, WebCorpus & DeepSeek-AI\\ \hline
Mistral 7B \protect\cite{Mistral7BO} & 7.42B & 1T (Estimated) & 8K & Yes & -- & 2023 & Open web data, books & Mistral AI \\ \hline
LLaMA 3 (70B) \protect\cite{lama3herdmodels} & 70B & 15T & 8K (standard), 128K (extended) & Yes & LLaMA 2 & 2024 & Public web corpora, code, academic texts, multilingual data & Meta (Facebook AI) \\ \hline
LLaMA 2 (13B) \protect\cite{Llama2} & 13B & 2T & 4K & Yes & LLaMA & 2023 & Public web data, curated datasets & Meta (Facebook AI) \\ \hline
Claude 3.5 Sonnet \protect\cite{anthropic_website} & Not publicly disclosed & Not publicly disclosed & 200K & No & Claude 3 Sonnet & 2024 & Proprietary, refined on large-scale data & Anthropic \\ \hline
Phi-2 \protect\cite{phi2_blog} & 2.7B & 1.4T & 2K & Yes & Phi-1.5 & 2023 & Proprietary, code-focused datasets & Microsoft \\ \hline
Vicuna-13B \protect\cite{vicuna_blog} & 13B & 1T & 2K & Yes & LLaMA & 2023 & OpenAI GPT 3.5 dataset, open web data & LMSYS \\ \hline
PaLM-2 \protect\cite{palm2technicalreport} & 340B & 3.6T & 8K & Yes & PaLM & 2023 & Web documents, books, code, math, conversational data & Google \\ \hline
BERT \protect\cite{DevlinBERT:Understanding} & 110M & 137B & 512 & Yes & -- & 2018 & BookCorpus, English Wikipedia & Google \\ \hline
DeBERTa \protect\cite{DeBERTa} & 1.5B & Not publicly disclosed & 512 & Yes & BERT & 2021 & BooksCorpus, Wikipedia, STORIES, Reddit content & Microsoft \\ 
\hline\hline
\end{tabular}
\label{model_comparison}
\end{sidewaystable}

\subsection{Large Language Model Meta AI 3}

Following the success of Large Language Model Meta AI (LLaMA) 2, Meta introduced LLaMA 3, which brings notable improvements in multi-turn reasoning and instruction following capabilities. This next-generation model builds upon its predecessor by integrating larger context windows, enhanced alignment mechanisms, and better optimization for dialogue-centric tasks. In the LLaMA 3 technical report \cite{lama3herdmodels}, Meta highlighted the model’s effectiveness on benchmarks, i.e., SQuAD, QuAC, and InfiniteBench, showing improved contextual understanding and coherence in multi-turn settings. The study also presented the model's ability to handle complex instructions and maintain relevance over extended dialogues. These capabilities make LLaMA 3 a strong candidate for ConvQA systems, especially in applications requiring tool use, structured memory, or long-form interaction support. Enhancing LLaMA 3-based conversational agents with Theory of Mind (ToM) was proposed in \cite{Llama3_2} to better align generated responses with human mental states, i.e, beliefs, desires, and intentions. By incorporating a LatentQA pipeline to interpret and steer internal model representations, the study demonstrates that ToM-informed alignment substantially improves contextual understanding and response quality in multi-turn conversations, thereby achieving up to 67\% win rates on benchmark datasets, i.e., CaSiNo and NegotiationToM. These results underscore the potential of LLaMA 3 models in building controllable, socially-aware, and cognitively aligned conversational systems.

Table \ref{model_comparison} delineates further details on the discussed models along with additional LLMs by highlighting their unique characteristics and expanded capabilities relevant to ConvQA.

\section{Datasets for Conversational Question Answering Systems}

Datasets play a crucial role in the development and evaluation of ConvQA systems. They provide the necessary resources for training and benchmarking models, thereby ensuring their ability to understand and generate contextually relevant responses. A well-constructed dataset enables the ConvQA system to model multi-turn interactions, resolve contextual dependencies, and handle real-world phenomena, i.e., ambiguity, topic shifts, and ellipsis. Furthermore, the diversity and structure of datasets significantly impact a model’s ability to generalize across domains and maintain consistency in long-range conversational flows. As ConvQA research advances, recent datasets have begun incorporating challenging features, i.e., topic switching, open-domain settings, and entity-linked knowledge grounding, thereby making them indispensable tools for building robust and adaptive systems. Here are some notable datasets commonly used in ConvQA research.

\subsection{Conversational Question Answering Dataset}
Conversational Question Answering (CoQA) dataset \cite{reddy2019coqa} contains 127,000+ questions with answers obtained from 8,000+ conversations. It is designed to assess a model's ability to understand and answer questions within the context of a dialogue. Each conversation is based on a passage from diverse domains, i.e., literature, middle/high school exams, and news articles. CoQA emphasizes the importance of context as questions often rely on previous conversational turns for accurate responses. A comparative analysis of  SQuAD 2.0, QuAC, and CoQA datasets was conducted in \cite{CoQA1} highlighting CoQA's unique features, i.e., multi-turn interactions and the inclusion of both extractive and abstractive answers. Using the BiDAF++ model, the study demonstrated that CoQA achieved high performance by surpassing existing baselines. Additionally, the research showed moderate improvements in performance through cross-dataset transfer with fine-tuning. Furthermore, the CoQA dataset was utilized in \cite{CoQA2} to evaluate their proposed prompt-guided copy mechanism which aims to improve the fluency and coherence of answers in ConvQA. By training on the CoQA dataset and applying tailored prompts, their method showed significant improvements in both extractive and generative answer accuracy. This demonstrated their method's effectiveness in handling the conversational nuances of the CoQA dataset.

\subsection{Question Answering in Context Dataset}
Question Answering in Context (QuAC) dataset \cite{Quac} includes 100,000+ questions and answers in 14,000+ conversations. It focuses on answering questions about Wikipedia articles in a conversational setting. This dataset is unique in its emphasis on information-seeking conversations, where the questions are not pre-specified but generated dynamically during the interaction. QuAC tests a ConvQA model's ability to generate contextually relevant and coherent answers over multiple conversational turns. The QuAC dataset was deployed in \cite{KTurn1} to evaluate a question rewriting model which improved the accuracy of extracting answer spans by rewriting ambiguous questions into clear context-independent questions, thereby enhancing overall QA performance. Moreover, the QuAC dataset was leveraged in \cite{EXCORD} to validate the ExCoRD framework, thereby demonstrating a significant improvement in QA performance by addressing conversational dependencies through consistency training. 

\subsection{Stanford Question Answering 2.0 Dataset}

Stanford Question Answering 2.0 (SQuAD 2.0) dataset \cite{Squad2.0} is an extension of SQuAD 1.1 dataset. It combines 100,000+ questions from the original dataset with 50,000+ unanswerable questions crafted adversarially. While SQuAD is not inherently conversational, it has been adapted for ConvQA tasks by considering the sequence of questions as a conversation. The challenge lies in distinguishing between answerable and unanswerable questions which often require deep contextual understanding. RoBERTa model was fine-tuned on SQuAD 2.0 in \cite{SQUAD1} by focusing on the challenges of unanswerable questions to assess model performance. The authors delineated that SQuAD 2.0's inclusion of unanswerable questions required models to demonstrate deeper reasoning. In addition, the SQuAD 2.0 dataset was employed in \cite{SQUAD3} to evaluate the generalizability of BERT-based models. The results revealed that while these models perform well, they often rely on heuristics rather than true comprehension, thereby highlighting limitations in the dataset's ability to test deep understanding.

\subsection{Conversational Answer Reformulation Dataset}

Conversational Answer Reformulation Dataset (CANARD) \cite{CANARD} consists of 40,000+ question-answer pairs derived from the QuAC dataset. The primary focus of CANARD is on rewriting conversational questions into stand-alone questions. This process involves rewriting questions that are dependent on the previous context to make them context-independent. The dataset is instrumental in training models to handle conversational dependencies and improve their ability to understand and generate contextually coherent responses in a dialogue setting. By transforming context-dependent questions into context-independent questions, the CANARD dataset enhances the ConvQA system's comprehension and answering capabilities in multi-turn conversations. The CANARD dataset was utilized in \cite{KTurn1} to train a question rewriting model which improved the clarity of follow-up questions in conversational contexts, thereby leading to enhanced performance in both retrieval and extractive QA tasks. Furthermore, the CANARD dataset was deployed in \cite{Zaib2023KeepingAnswering} to train a ConvQA system that generates structured representations instead of rewriting questions. This approach improved the model's ability to handle ambiguous questions by leveraging context and question entities from previous conversational turns.

\begin{sidewaystable}[htbp]
\normalsize
\renewcommand{\arraystretch}{2}
\setlength{\tabcolsep}{1.0pt}
\centering
\caption{Comparison of conversational question answering datasets (CoQA, QuAC, SQuAD 2.0, CANARD, QReCC, and TopiOCQA).}
\label{tab:dataset_comparison}

\begin{tabular}{>{\centering\arraybackslash}m{2.2cm}|>{\centering\arraybackslash}m{4.5cm}|>{\centering\arraybackslash}m{2cm}|>{\centering\arraybackslash}m{2.8cm}|>{\centering\arraybackslash}m{4cm}|>{\centering\arraybackslash}m{8cm}}
\hline\hline
\textbf{Dataset} & \textbf{Source} & \textbf{Size / Number of Questions} & \textbf{Number of Conversations} & \textbf{Question Type} & \textbf{Special Features} \\ \hline
CoQA \cite{reddy2019coqa} & Children's stories, literature, mid/high school exams, news, Wikipedia, Reddit, science. & 127K & 8K & Conversational, free-form text, unanswerable & Multi-turn, text-based, dialog acts, pragmatic reasoning, coreference resolution, diverse domains. \\ \hline
QuAC \cite{Quac} & Wikipedia articles & 100K & 14K & Information-seeking, open-ended & Multi-turn, text-based, dialog acts, unanswerable questions, context-driven. \\ \hline
SQuAD 2.0 \cite{Squad2.0} & Wikipedia articles & 129K & N/A (single-turn questions) & Fact-based, answerable and unanswerable questions &  Unanswerable questions to test model's ability to detect when no answer is present, context-driven, challenging distractors. \\ \hline
CANARD \cite{CANARD} & QuAC dataset & 40K & 5K & Query rewriting, context-dependent & Focuses on question rewriting, contextually aware, requires an understanding of conversational history. \\ \hline
QReCC  \cite{QRECC} & QuAC, NQ, TREC CAsT datasets & 80K & 13.6K & Query rewriting, conversational, context-dependent & Multi-turn, open-domain, focuses on question rewriting, passage retrieval, and reading comprehension. \\ \hline
%SimQuAC \cite{2024_05} & Wikipedia articles from QuAC dataset & 4005 & 342 & Synthetic, information-seeking, open-ended & LLM Simulated, multi-turn, text-based, dialog acts, unanswerable questions, and context-driven. \\ \hline
TopiOCQA \cite{2022_01Dataset} & Wikipedia Articles & 50.5K & 3920 & Conversational, open-domain, information-seeking, free-form	& Multi-turn, open-domain, topic-switching, free-form answers, complex coreference, and 4 topics per conversation on average.
\\ \hline\hline

\end{tabular}
\label{tab:dataset_comparison}
\end{sidewaystable}

\subsection{Question Rewriting in Conversational Context Dataset}

Question Rewriting in Conversational Context (QReCC) dataset \cite{QRECC} is a large-scale, open-domain ConvQA dataset designed to extend traditional QA tasks into a conversational context. It consists of 13.6K conversations with 80K question-answer pairs where questions are often context-dependent, thereby requiring rewriting to be understood outside the dialogue. The dataset uniquely combines question rewriting, passage retrieval, and reading comprehension sub-tasks, thereby making it a comprehensive resource for developing and evaluating end-to-end conversational QA models. The QReCC dataset was utilized in \cite{EntireHistory1} to train and evaluate their ConvQA system. Specifically, the QReCC dataset provided a comprehensive framework for assessing the entire pipeline of their system, including question rewriting, passage retrieval, and answer generation. Moreover, the QReCC dataset was utilized in \cite{QRECC2} to retrieve relevant information from large text collections and generate accurate answers. The task was divided into question rewriting, passage retrieval, and answer generation. The dataset was also extended by adding alternative correct answers produced by participant systems to enhance model evaluation.

\subsection{Topic Switching in Open-domain Conversational Question Answering Dataset} 

Topic Switching in Open-domain Conversational Question Answering (TopiOCQA) dataset \cite{2022_01Dataset} is a large-scale open-domain ConvQA dataset that reflects topic switching, common in real-world information-seeking conversations. Unlike earlier datasets restricted to single topics or requiring reference passages, TopiOCQA simulates dynamic user interactions across related topics, with conversations averaging 13 turns and covering 4 distinct Wikipedia documents. It contains 3,920 conversations and over 50,000 question-answer pairs, where the questioner sees only metadata (titles and section headers), while the answerer navigates full documents via hyperlinks. This setting introduces challenges, i.e., coreference resolution, ambiguous questions, and retrieval across shifting topics.  The TopiOCQA dataset was used in \cite{2024_7Topic} to train and evaluate question rewriting models for multi-turn ConvQA with topic shifts. The authors fine-tuned their models on over 45,000 TopiOCQA instances, thereby emphasizing non-initial turns to capture contextual dependencies. They tested the rewritten queries using both sparse and dense retrievers to measure retrieval effectiveness. This setup enabled robust assessment of context-aware query rewriting in complex and document-grounded conversations. Furthermore, The TopiOCQA dataset was utilized in \cite{2024_08Topic} to evaluate how well LLMs handle multi-turn ConvQA interactions. The dataset, which features questions contextualized within dialogues, helped assess both the model's ability to extract answers from relevant context and to disregard distracting context. Special prompting was applied to indicate conversational turns using a <SEP> token, thereby ensuring the model was aware of the conversation structure.

Table \ref{tab:dataset_comparison} delineates further details on the discussed datasets by highlighting their unique characteristics, structural properties, and distinguishing features.

\section{Open Research Directions}

\begin{figure}[!t]
    \centering
    \includegraphics[width=0.7\textwidth]{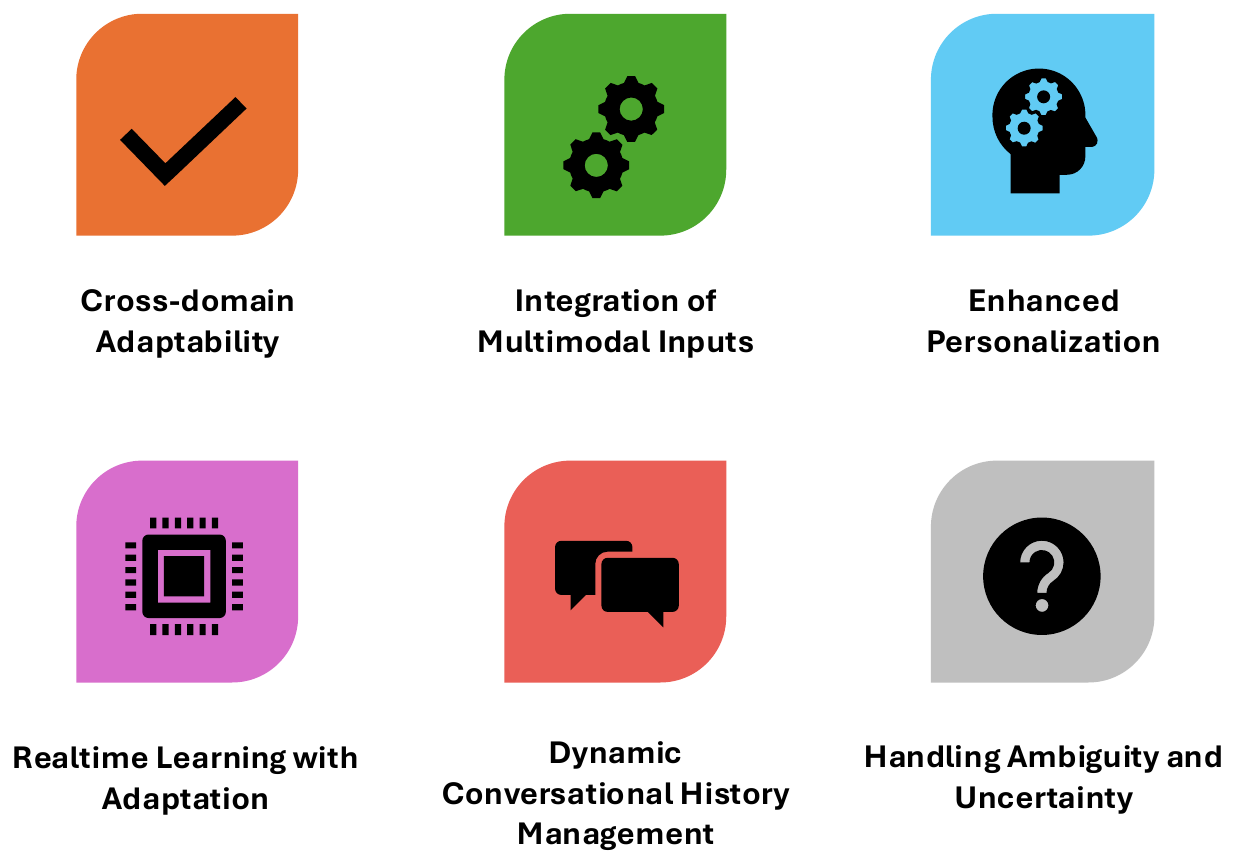} % Adjust the width as needed
\vspace{10pt} 
    \captionsetup{justification=centering}
    \caption{Open research directions in conversational question answering -- Key areas include cross-domain adaptability, integration of multimodal inputs, enhanced personalization, real-time learning with adaptation, dynamic conversational history management, and handling ambiguity and uncertainty.}
    \label{fig:odc}
\end{figure}

As ConvQA systems evolve, these systems encounter unforeseen challenges and opportunities that demand innovative approaches. The future direction of ConvQA will be shaped by research that addresses the challenges of maintaining contextual coherence.
This involves the integration of external knowledge and improving user experience across numerous domains. As delineated in Figure \ref{fig:odc}, the following open research directions delineate the critical areas where future research can significantly contribute to the advancement of more sophisticated ConvQA systems.

\subsection{Cross-domain Adaptability}

Cross-domain adaptability is an increasingly important goal for ConvQA systems, as users often shift between topics and domains during real-world conversations in areas, i.e., customer service, healthcare, or education. Most existing ConvQA models are trained within a specific domain, which limits their ability to handle diverse or unseen queries. To address this limitation, future research should focus on developing models that can generalize across multiple domains without the need for extensive retraining. Techniques, i.e., TL and domain adaptation, help models apply knowledge learned from one domain to another. Recent advancements in instruction tuning, in-context learning, and prompt-based adaptation have also enabled LLMs to perform well in unfamiliar domains using minimal supervision. Modular fine-tuning strategies, i.e., AdapterFusion and LoRA allow domain-specific adjustments without changing the core model, thereby supporting scalability. Additionally, metadata-aware retrieval, multi-task learning, and prompt routing mechanisms can further improve domain sensitivity and contextual accuracy. For robust evaluation, ConvQA systems should be tested on diverse, multi-domain datasets that reflect realistic and dynamic conversations. Synthetic data generation and domain-interleaved training can help simulate such settings. Finally, incorporating user feedback and RL from interaction histories can enable ConvQA systems to adapt continuously and improve performance over time across varying domains.

\subsection{Integration of Multimodal Inputs}

Although text-based inputs have been the main focus of current ConvQA systems, future systems should be able to process and integrate multimodal inputs, including images, videos, and audio. This would enable users to ask questions that involve different forms of media, i.e., asking questions about an image or video clip. For instance, a user could submit an image of a product and request specific information about it, thereby requiring the ConvQA system to comprehend and analyze both the visual content and the associated text of the user query. To develop such capabilities, it would be essential to develop ConvQA systems that can integrate information from multiple sources. This would ensure that ConvQA systems can deliver precise and contextually appropriate answers. Incorporating this approach would greatly improve the level of interactivity and effectiveness of the ConvQA systems in various domains, i.e., e-commerce, healthcare, and education. Recent breakthroughs in Vision Language Pretraining (VLP) and unified multimodal transformers, i.e., Gemini and Contrastive Language Image Pretraining (CLIP) by Open AI, have laid the groundwork for such systems. These models demonstrate the feasibility of aligning visual and textual representations within a shared embedding space. Future ConvQA research should explore fusion strategies, cross-modal attention, and alignment techniques to maintain consistency and coherence across modalities. Evaluation benchmarks should also evolve to assess multi-turn, multimodal reasoning capabilities in real world applications.

\subsection{Enhance Personalization}

Personalization is a critical frontier in advancing ConvQA systems, aiming to deliver more engaging, context-aware, and user-centric interactions. Existing ConvQA systems typically process each interaction independently, disregarding important contextual signals, i.e., user preferences, historical conversational patterns, and long-term goals. This stateless design limits the ConvQA system’s ability to adapt and provide responses that resonate with individual users’ intents or conversational styles. To overcome this limitation, future ConvQA systems should incorporate mechanisms that dynamically learn and retain user-specific details, i.e., frequently asked topics, preferred response formats, domain expertise levels, emotional tone preferences, and query intent patterns. Such personalization could be facilitated through techniques, i.e., long-term memory modeling, RL from user feedback, and the integration of user embeddings or context-aware language adaptation layers. However, this level of personalization introduces significant ethical and technical challenges. Foremost among these is the need to uphold data privacy, transparency, and fairness. ConvQA systems must be designed with mechanisms for consent-driven data collection, anonymization, and on-device personalization to reduce the risk of data leakage or misuse. Moreover, personalization algorithms must be monitored for potential biases that could result from overfitting to individual user traits or reinforcing harmful patterns.

\subsection{Real-time Learning with Adaptation}

Recent advancements in ConvQA systems highlight the critical importance of real-time learning and adaptation, particularly in rapidly evolving and information intensive environments. Unlike traditional systems that rely on static or periodically updated knowledge bases, modern ConvQA architectures now incorporate advanced real-time learning methodologies, i.e., RL, AL, CL, online machine learning, and meta-learning. The integration of RAG pipelines and enhanced long-context handling techniques has further enabled seamless updates to knowledge bases and adaptive response strategies during interactions. State-of-the-art models, i.e., GPT, Gemini, LLaMA, Mistral, and Claude, have demonstrated remarkable capabilities in dynamic contextual adaptation, thereby enhancing the precision and relevance of responses. Complementary to these, self-supervised learning approaches reduce dependency on annotated data by leveraging unlabeled conversational inputs, while user-feedback-driven fine-tuning supports situational calibration in real time. Furthermore, federated learning methods are being explored to support decentralized adaptation with privacy preservation, and continual learning pipelines help maintain performance over prolonged use without catastrophic forgetting. Collectively, these advancements are steering ConvQA systems toward becoming more adaptive, personalized, and robust in delivering timely and contextually aligned answers in real-world applications.

\subsection{Dynamic Conversational History Management}

Effectively managing extensive conversation histories presents a substantial challenge in ConvQA, particularly in the context of lengthy multi-turn interactions. Existing ConvQA systems frequently struggle to maintain context during lengthy conversations, thereby leading to responses that are irrelevant or incoherent. Further research should prioritize the exploration of dynamic history management techniques that can selectively retain, summarize, or remove parts of the conversational history based on their relevance to the current conversation. This may involve developing relevance-based filtering mechanisms that prioritize important conversational turns, along with summarization techniques that transform lengthy conversational turns into summarized representations. Dynamic history management would enable the ConvQA systems to maintain coherence and provide contextually appropriate answers, even during prolonged interactions. In addition to summarization and filtering, future ConvQA systems may also leverage RL and attention-based strategies to dynamically adjust history selection policies during inference. Techniques, i.e., memory compression, entity tracking, and semantic clustering, can further aid in reducing redundancy while preserving essential contextual cues. Incorporating external memory structures or graph-based conversation representations may also improve the ConvQA model’s ability to trace back relevant information efficiently. Moreover, real-time history adaptation based on evolving user goals or topic shifts remains an open area for exploration. These advancements are critical for scaling ConvQA systems in real-world applications, i.e., virtual tutoring, healthcare, and multi-session customer service, wherein sustained context modeling is essential.

\subsection{Handling Ambiguity and Uncertainty}

Users often pose ambiguous or uncertain queries to ConvQA systems in real-life scenarios, thereby making it challenging for ConvQA systems to provide accurate answers. Current ConvQA systems typically aim to offer a single answer, which can be misleading in cases where the query is unclear or the information is not definitive. Future ConvQA systems should be optimized to better manage ambiguity and uncertainty by providing multiple plausible responses, along with confidence scores and explanations. This approach would assist users in comprehending a range of potential responses and making well-informed decisions based on the information provided. Handling ambiguity and uncertainty is particularly important in high-stakes applications, i.e., healthcare or legal advice, where providing multiple options or clarifying uncertainties can significantly impact the outcome of the interaction. To achieve this, ConvQA systems could leverage probabilistic reasoning, uncertainty-aware language modeling, and contradiction detection techniques to flag ambiguous queries and generate diversified answers. Incorporating interactive clarification mechanisms, where the system actively asks follow-up questions to resolve ambiguity, can also enhance conversational robustness. Furthermore, aligning ConvQA system confidence with human interpretability ensures users can better assess the reliability of the provided answers, thereby improving trust and user satisfaction in real-world deployments. Recent advancements in LLMs also introduce self-reflection and selective answering capabilities, thereby enabling the system to acknowledge uncertainty, reject low-confidence answers, or verify responses through internal consistency checks.

These open research directions highlight the need for more adaptive, context-aware, and user-centric ConvQA systems. By addressing these challenges, future researchers can develop next-generation ConvQA systems that are more robust, versatile, and capable of handling the complexities of real-world conversations.

\section{Conclusion}

This survey has provided a comprehensive exploration of the state-of-the-art in ConvQA, covering key components, advanced techniques, LLMs, and datasets that have shaped this rapidly evolving field. As ConvQA systems continue to progress, the integration of key components, i.e., history selection, question understanding, and answer prediction remains paramount in ensuring coherence and relevance in multi-turn conversations. The survey also highlighted the significant impact of advanced machine learning techniques, i.e., RL, KD, CL, AL, and TL, in enhancing the performance and accuracy of ConvQA models. Moreover, the paper delved into the pivotal role of LLMs, including but not limited to, RoBERTa, GPT-4, Gemini 2.0 Flash, Mistral 7B, and LLaMA 3 which have set new benchmarks in ConvQA by leveraging vast amounts of data and sophisticated architectures. The detailed discussion on various datasets underscored their importance in training and evaluating ConvQA systems by facilitating the development of more robust and contextually aware models. Looking forward, the future of ConvQA holds immense potential, with key open research directions identified in this survey, i.e., cross-domain adaptability, integration of multimodal inputs, enhanced personalization, real-time learning with adaptation, dynamic conversational history management, and handling ambiguity and uncertainty. The pursuit of real-time learning and dynamic history management, along with improved handling of ambiguity and uncertainty, will be crucial in addressing the complex challenges that arise in real-world conversational contexts. As the field advances, the insights provided in this survey aim to guide researchers and practitioners in navigating the intricacies of ConvQA by fostering innovation and paving the way for more sophisticated and user-centric conversational systems. The continued evolution of ConvQA will be driven by these emerging trends and technologies. These advancements promise to transform how humans interact with machines by making interactions more natural, intuitive, and effective across various domains.

\vspace{20pt}
\noindent\textbf{Funding}: The research-at-hand is supported by the International Macquarie University Research Excellence Scholarship (Allocation Number: 20257342).

\vspace{20pt}
\noindent\textbf{Authors' Contributions}: M.M.P. conceptualized the research, carried out the literature review, and primarily wrote the initial draft. A.M. and Q.Z.S. were also involved in the conceptualization of the research, and provided critical supervision and salient guidance pertinent to the scope and the structure of the paper. They also contributed considerably in refining the final version. K.E.W., F.I., and M.T. assisted in reviewing relevant research literature and contributed to the specific sections of the initial draft. All authors reviewed and approved the final manuscript.

\bibliography{sn-bibliography}% common bib file

\end{document}